\pdfoutput=1

\documentclass[11pt]{article}

\usepackage[]{acl}

\usepackage{times}
\usepackage{latexsym}
\usepackage{multirow}
\usepackage{graphicx}
\usepackage{makecell}
\usepackage{amssymb}
\usepackage{pifont}
\usepackage{enumitem}
\usepackage{enumitem}

\usepackage[T1]{fontenc}

\usepackage[utf8]{inputenc}

\usepackage{microtype}

\usepackage{inconsolata}
\usepackage{float}
\restylefloat{table}
\usepackage{graphicx}
\usepackage{subcaption}
\title{Leveraging Large Language Models for Learning Complex Legal Concepts through Storytelling}

\author{Hang Jiang$^{\dagger\varnothing}$, Xiajie Zhang$^{\dagger}$, Robert Mahari$^{\dagger\diamondsuit}$, Daniel Kessler$^{\dagger\varnothing}$, Eric Ma$^{\Box}$, \\
\bf Tal August$^{\infty}$, Irene Li$^{\neg\dashv}$, Alex `Sandy' Pentland$^{\dagger}$, Yoon Kim$^{\dagger}$, Deb Roy$^{\dagger\varnothing}$, Jad Kabbara$^{\dagger\varnothing}$\\
  $^{\dagger}$Massachusetts Institute of Technology, $^{\varnothing}$MIT Center for Constructive Communication\\ $^{\diamondsuit}$Harvard Law School,
  $^{\Box}$University of Virginia School of Law\\
  $^{\infty}$Allen Institute for AI, $^{\neg}$University of Tokyo, $^{\dashv}$Smartor.me \\
}

\begin{document}
\maketitle
\begin{abstract}

Making legal knowledge accessible to non-experts is crucial for enhancing general legal literacy and encouraging civic participation in democracy. However, legal documents are often challenging to understand for people without legal backgrounds. In this paper, we present a novel application of large language models (LLMs) in legal education to help non-experts learn intricate legal concepts through storytelling, an effective pedagogical tool in conveying complex and abstract concepts. We also introduce a new dataset \textsc{LegalStories}, which consists of 294 complex legal doctrines, each accompanied by a story and a set of multiple-choice questions generated by LLMs. To construct the dataset, we experiment with various LLMs to generate legal stories explaining these concepts. Furthermore, we use an expert-in-the-loop approach to iteratively design multiple-choice questions. Then, we evaluate the effectiveness of storytelling with LLMs through randomized controlled trials (RCTs) with legal novices on 10 samples from the dataset. We find that LLM-generated stories enhance comprehension of legal concepts and interest in law among non-native speakers compared to only definitions. Moreover, stories consistently help participants relate legal concepts to their lives. Finally, we find that learning with stories shows a higher retention rate for non-native speakers in the follow-up assessment. Our work has strong implications for using LLMs in promoting teaching and learning in the legal field and beyond.

\end{abstract}

\section{Introduction}


Often individuals find themselves in certain high-stakes situations where they have to educate themselves on novel concepts such as new policies before voting, mortgage terms when buying a house or legal principles relevant to an ongoing lawsuit. Unfamiliar terms and nuanced use of language in these contexts can make it challenging for non-experts to make informed decisions, to have equal access to justice, or to participate in civic discourse and democracy. We present this work as a step towards enhancing general legal literacy, bridging the gap between non-experts and experts and promoting constructive and civic discourse. 


Storytelling is an important medium to communicate science to non-experts \cite{dahlstrom2014using,martinez2017finding} and teach professional knowledge to beginners \cite{abrahamson1998storytelling,davidhizar2003storytelling,gallagher2011search}. In legal contexts, storytelling has been used extensively to teach abstract legal concepts such as ethics \cite{menkel1999winning}, and has proven effective at explaining complex legal concepts such as legal mediation to the general public \cite{Capuano2014ASL}. 
However, the scalable implementation of legal storytelling education is severely limited by the high costs associated with legal experts.

Large language models (LLMs) and their impressive text generation abilities have facilitated high-quality automated explanations and stories. Recent efforts \cite{huang-etal-2021-definition,murthy2021towards,murthy-etal-2022-accord,august-etal-2022-generating} have leveraged LLMs to generate accessible explanations of scientific or medical concepts for diverse audiences. \citet{savelka2023explaining} used GPT-4 to generate explanations for legal concepts from statutory provisions. However, to the best of our knowledge, previous work has not: (1) used LLM-generated stories as a medium to explain complex concepts, especially in the under-explored legal domain, (2) generated and refined (via expert feedback) questions for the assessment of concept comprehension, nor (3) validated the effectiveness of LLM-generated stories in enhancing comprehension among non-experts.

In this work, we explore a novel application of LLMs that focuses on the use of generated stories and questions to facilitate the learning and assessment of legal concept understanding. We use a human-in-the-loop pipeline that combines LLM and expert input to generate stories and multiple-choice questions. We loop in both Prolific workers and legal experts\footnote{This paper uses the term ``legal experts'' to refer to two people who have graduated with JD degrees or have made substantial progress towards earning JD degrees.} to ensure that the LLM-generated content is of high-quality. Our pipeline presents a holistic approach to LLMs' application in the legal education domain, where both the learning intervention (stories) and assessment (reading comprehension questions) are generated and evaluated. By providing a reusable dataset and promising experiment results, our work has strong implications for the broader use of LLMs to enhance teaching and learning and to improve general legal literacy. Our contributions are as follows:

\begin{itemize}[itemsep=1pt, topsep=1pt]
    \item We create a novel legal education dataset, \textsc{LegalStories}, which presents legal concepts with their definitions, LLM-generated stories and questions, and human annotations for future NLP and legal education research\footnote{Both the code and data are available at this repository: \url{https://github.com/hjian42/LegalStories}.}.
    \item We provide extensive comparisons of three LLMs, namely, LLaMA 2, GPT-3.5, and GPT-4, to generate legal stories and questions with both automatic and human evaluations. 
    \item We conduct RCTs with both native and non-native English speakers to learn legal concepts, demonstrating that LLM-generated stories improve concept comprehension and interest in law among non-native speakers compared to Wikipedia definitions. We also find that LLM-generated stories consistently help both native and non-native participants in relating legal concepts to their personal lives.
\end{itemize}

\section{Related Work}

\paragraph{Legal NLP \& Accessible Language}
Legal language is complex and nuanced, and this creates a challenge for non-experts navigating legal processes~\cite{benson1984end}.
This challenge has prompted research into using computational tools to improve legal reading comprehension~\cite{curtotti2013right}. 
Making legal jargon more accessible represents an impactful application of legal NLP that promises to broaden access to justice~\cite{mahari-etal-2023-law}. Previous work has focused on legal text simplification \cite{collantes2015simpatico,garimella-etal-2022-text,cemri2022unsupervised}, legal summarization \cite{farzindar-lapalme-2004-legal,manor-li-2019-plain}, and question answering \cite{khazaeli-etal-2021-free,zhong2020jec,martinez2023survey} to make legal language more accessible. LLMs \cite{zhang2023unleashing} have also been used to improve legal access such as ChatLaw \cite{cui2023chatlaw}. Recently, \citet{savelka2023explaining} used GPT-4 to explain legal concepts in statutory provisions. However, none of the previous legal NLP work has combined LLMs and storytelling--a widely-used technique in education and communication--as a device to bridge legal experts and non-experts. 





\paragraph{Storytelling in Education and NLP}
\textit{Stories}, which we use interchangeably with the term \textit{narratives}, are sequential depictions of actions and events \cite{abbott2020cambridge}. They are an effective technique for pedagogy \cite{busselle2009measuring,busselle2008fictionality,rapaport1989deictic}. Stories effectively illustrate complex concepts in various fields like math, science, and law, with law education heavily relying on ``fact-patterns'' \cite{papadimitriou2003mythematics}. Specifically, second-person narratives are known to elicit particularly strong emotional and aesthetic-reflexive involvement from readers \cite{mildorf2016reconsidering}, especially when combined with more advanced narrative techniques and emotive verbs \cite{rembowska2022enactive}. Automatic story generation is a long-standing task in NLP. Before LLMs, the best story generation models, even those using transformer architectures, struggled to create coherent stories with well-defined characters and story-lines \cite{alabdulkarim-etal-2021-automatic}. Recent progress in LLMs such as ChatGPT have opened up new possibilities with storytelling, showing significantly higher quality compared to other story generation models \cite{xie-etal-2023-next,zimmerman2022future,xu-etal-2020-megatron}. However, little work has explored the use of LLM-generated stories in education. \citet{valentini2023automatic} experimented with LLMs to generate age-appropriate stories for children. Our work is the first to use LLM-generated stories to help non-experts in learning legal knowledge.

\begin{figure*}[t]
\centering
\includegraphics[clip, trim = 0px 0px 0px 00px,width=0.99\linewidth]{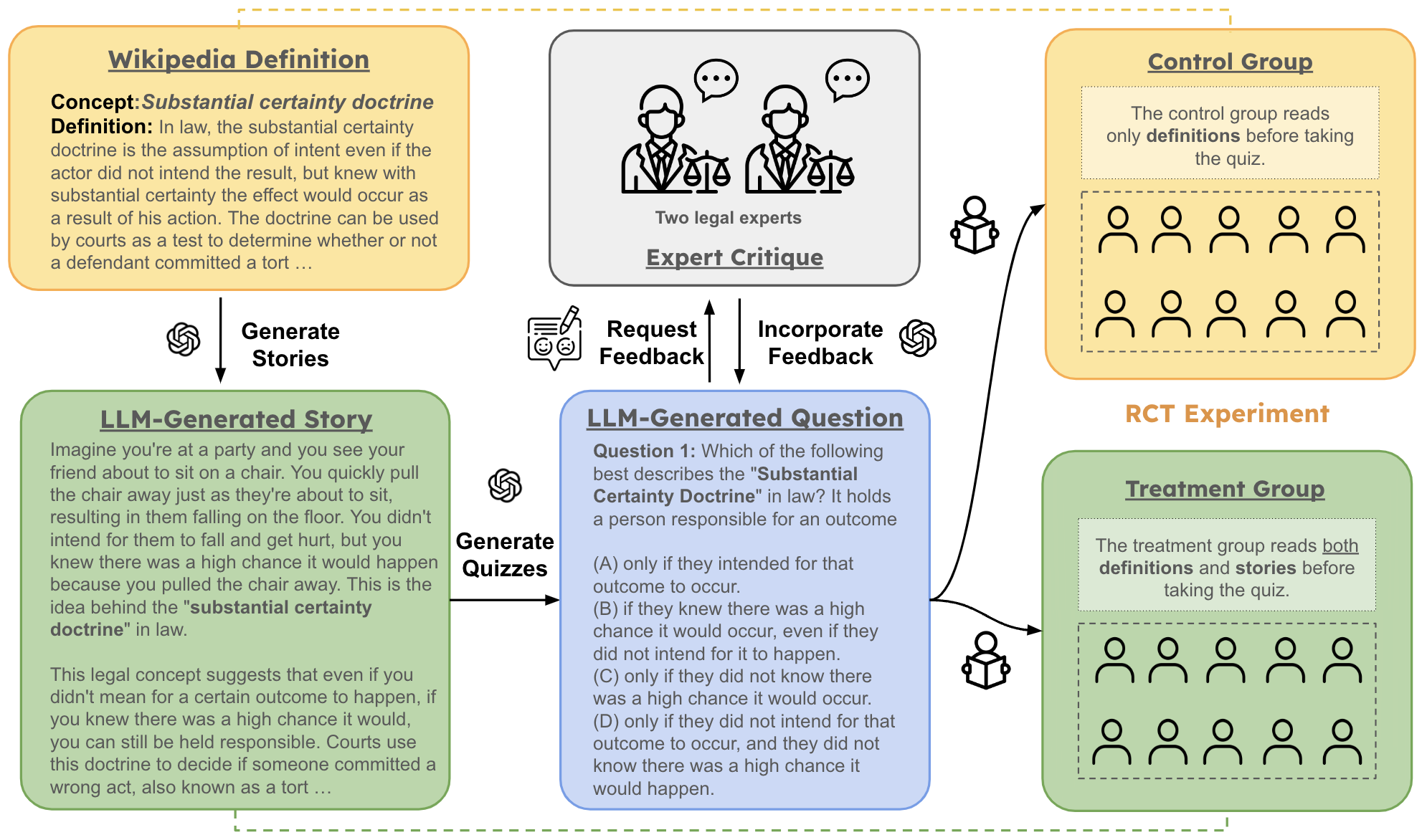}
\caption{Illustration of the expert-in-the-loop pipeline. The left section demonstrates the procedure to produce an LLM-generated story from the concept. The lower section in the center shows how we use both the definition and story as input to produce LLM-generated reading comprehension (RC) questions. The center upper section shows that we first collect expert feedback on questions and regenerate questions with expert advice. The right section outlines the RCT experiment to see if LLM-generated stories improve comprehension in legal concepts.
}
\label{fig:framework}
\vspace{-2mm}
\end{figure*}

\paragraph{Educational Question Generation}
Generating educational questions or quizzes is important for educators to increase engagement, test reading comprehension, and improve learners' knowledge retention \cite{al2023review}. In the era of modern NLP, question generation (QG) was first tackled with seq2seq 
\cite{yuan-etal-2017-machine,zhou2018neural} and Transformer-based models 
\cite{narayan2020qurious,bao2020unilmv2}. Previous QG work tackled selecting question-worthy content 
\cite{liu2020asking,steuer2021not}, modeling question types 
\cite{duan-etal-2017-question,sun-etal-2018-answer,sugawara2018makes,kang-etal-2019-know,zhou-etal-2019-question}
, or focused on specific interrogative words such as what, how, etc. \cite{ghanem-etal-2022-question}. A subset of work focuses on the needs of students and teachers to tackle question generation to make them more accessible, adaptive, and educational \cite{le2014automatic,wang2018qg,chen2018learningq,kurdi2020systematic,srivastava-goodman-2021-question,wang2022towards,leite2023towards,laban-etal-2022-quiz}.
Recently, LLMs have been widely used to generate questions  \cite{wang2022towards,gabajiwala2022quiz,muse2023pre,kasneci2023chatgpt,liang2023prompting,tran2023generating,lu2023readingquizmaker} 
in different domains such as language learning \cite{xiao-etal-2023-evaluating}, commonsense knowledge \cite{rathod-etal-2022-educational}, coding \cite{sarsa2022automatic,macneil2023implications}, and science \cite{bulathwela2023scalable}. However, prior research primarily evaluated generated questions via automatic metrics or simple human ratings. Differently from previous work, \citet{steuer2022educational} evaluated the usability of generated questions in a practical learning setting and showed that asking non-native learners to finish questions after reading helps them understand science texts. Our work instead uses LLM-generated questions as a tool to assess the effectiveness of LLM-generated stories in legal concept learning.

\section{\textsc{LegalStories} Dataset}
\label{sec:dataset_generation}

In Figure~\ref{fig:framework}, we present a pipeline and demonstrate how we apply it to curate a new dataset. It consists of three components: Story Generation, Question Generation, and Expert Critique. The following paragraphs illustrate how we generate stories and questions from legal doctrine definitions and refine questions with expert feedback.

\subsection{Doctrine Definitions from Wikipedia}
We collect 294 legal doctrines in English from the ``legal doctrines and principles'' page on Wikipedia\footnote{\url{https://en.wikipedia.org/wiki/Category:Legal_doctrines_and_principles}}. A legal doctrine is a systematic framework or set of rules and procedures that evolves through legal precedent, particularly in common law systems. It serves as a guide for determining judgments in legal cases. The development of a legal doctrine often begins when a judge makes a ruling that involves a specific process, which is then outlined and applied to the case. If this process is consistently used by other judges in similar situations and gains acceptance, it may become an established and widely accepted method for deciding similar cases in the future. A legal doctrine involves complex legal concepts whose definitions usually contain one or multiple legal terms, therefore difficult to comprehend to legal novices. We use the introduction paragraph as the definition for each legal doctrine. In our work, we focus on evaluating 101 out of 294 concepts whose definition length is between 100 and 200 words to ensure that they could be mapped to a medium-length story with around 500 words. We did not choose concepts whose definitions are too short or too long, because short definitions are not fair for participants in the control group to understand the concept and long definitions tend to have details missed by LLMs due to story length constraint. The mean and median lengths of the 101 concept definitions are 140.1 and 136.0 words.

\subsection{Story and Question Generation}
In this section, we describe the procedure to generate explanatory stories of legal concepts and three types of reading comprehension questions. We experiment with LLaMA 2 \cite{touvron2023llama}, GPT-3.5\footnote{\url{https://openai.com/index/chatgpt}} and GPT-4\footnote{\url{https://openai.com/index/gpt-4-research}} \cite{achiam2023gpt} given that they are among the state-of-the-art chat-based LLMs. See Appendix \ref{appendix:models} for details regarding the models and prompts.


\subsubsection{Story Generation}

As illustrated in the leftmost yellow and green boxes of Figure \ref{fig:framework}, we generate legal stories based on corresponding concepts and definitions. We find that a simple prompt (see Appendix \ref{appendix:prompt_story}) is good enough to generate legal stories. We limit the story length to 500 words because the definitions are between 100 and 200 words and lengthy content tends to overwhelm the readers. For the twenty sampled concepts, the mean and standard deviation of the definition lengths are $152.0\pm 31.0$ words. For the corresponding twenty stories used for human evaluation, the mean and standard deviation of their lengths are $316.8\pm 51.6$ for GPT-4, $327.0\pm 50.7$ for GPT-3.5, and $250.5\pm 89.9$ for LLaMA 2.


\subsubsection{Question Generation}
Prior pedagogical research has highlighted different aspects of cognitive learning: remembering, understanding, applying, evaluating, analyzing, and creating \cite{adams2015bloom}. Inspired by this framework, we create three question types which are suitable for assessing learners' understanding of concepts. In these three cases, the model is asked to generate a multiple-choice question with a suggested answer and explanation, with each type assessing a certain kind of understanding as follows:
\begin{itemize}[itemsep=1pt, topsep=0pt]
    \item \textbf{Concept question} for definition \textit{understanding}: Here, the task requires the reader to pick the most precise description of a concept.
    \item \textbf{Prediction question} for \textit{applying} the concept to scenarios: Here, the task involves asking the reader to forecast the outcome of a hypothetical situation that is related to the concept.
    \item \textbf{Limitation question} for \textit{evaluating} and \textit{analyzing} the concept's shortcomings: Here, the task requires the reader to identify a limitation or exception to the corresponding concept.
\end{itemize}
As depicted in the blue box on the lower center of Figure \ref{fig:framework}, we condition the question generation on the corresponding concept, definition, and story. The exact prompts are presented in Appendix \ref{appendix:prompts}.

\paragraph{Question Refinement with Expertise} As outlined in the central gray box at the top of Figure \ref{fig:framework}, we recruit two legal experts to read the concepts and stories, answer the questions, and provide critiques. This step aims to check (1) whether the quality of the generated questions is good, (2) whether the answers along with the explanation suggested by LLMs are correct and (3) whether they have suggestions to improve these questions or explanations. After completion, we simply ask them to provide suggestions and use these suggestions to prompt LLMs to improve the content. To implement this, we use a simple prompt which asks the model to generate new questions (see Appendix \ref{appendix:expert_feedback}).


\section{Evaluation}
\label{sec:dataset_evaluation}
A two-fold evaluation is carried out as follows: (a) an evaluation to determine the quality of the generated stories relating to doctrines, and (b) an evaluation of the generated questions and answers, as well as their efficacy in assessing comprehension.

\setlength{\extrarowheight}{3pt}
\begin{table*}[ht!]
    \centering
    \footnotesize
    \resizebox{0.95\textwidth}{!}{%
    \begin{tabular}{cccccccccc}
        {\textbf{Model}}  & 
        {\textbf{RoD}}  &
        {\textbf{RoS}} & 
        {\textbf{Relevant}} & 
        {\textbf{Redundant}} &
        {\textbf{Cohesive}} & 
        {\textbf{Complete}}& 
        {\textbf{Factual}}& 
        {\textbf{Likeable}}& 
        {\textbf{Believable}} \\
         \Xhline{3\arrayrulewidth}
        \multicolumn{10}{c}{\textit{101 Concepts (Mean\textsubscript{STD})}} \\ 
        GPT-4 & $3.95_{1.04}$ & $4.66_{0.60}$ & $4.56_{0.71}$ & $4.00_{1.26}$ & $4.63_{0.62}$ & $4.57_{0.67}$ & $4.56_{0.69}$ & $4.36_{0.81}$ & $4.54_{0.74}$ \\
        \Xhline{3\arrayrulewidth}
        \multicolumn{10}{c}{\textit{Sampled 20 Concepts (Mean\textsubscript{STD})}} \\ 
        GPT-4 & $\mathbf{3.98}_{1.07}$ & $\mathbf{4.70}_{0.46}$ & $\mathbf{4.52}_{0.68}$ & $3.78_{1.29}$ & $\mathbf{4.57}_{0.56}$ & $\mathbf{4.58}_{0.53}$ & $\mathbf{4.52}_{0.62}$ & $\mathbf{4.42}_{0.79}$ & $\mathbf{4.48}_{0.70}$ \\
        GPT-3.5 & $3.30_{1.01}$ & $4.35_{0.68}$ & $4.20_{0.78}$ & $3.72_{0.80}$ & $4.30_{0.74}$ & $4.03_{0.78}$ & $4.12_{0.69}$ & $4.10_{0.95}$ & $4.13_{0.65}$ \\
        LLaMA 2 & $3.72_{1.15}$ & $4.35_{0.86}$ & $4.40_{0.85}$ & $3.92_{1.33}$ & $4.38_{0.83}$ & $4.15_{1.12}$ & $4.10_{1.17}$ & $4.20_{1.04}$ & $4.35_{0.94}$ \\

    \end{tabular}
    }
    \caption{Human evaluation results of LLM-generated legal stories. The upper section contains scores for GPT-4 on the complete 101 legal concepts. The lower section contains scores for GPT-4, GPT-3.5, and LLaMA 2 on a subset of 20 legal concepts. RoD and RoS indicates the readability of the definition and the story respectively.}
        \label{tab:story_human_eval}
\end{table*}

\subsection{Story Evaluation}
\subsubsection{Human Evaluation}
For each legal concept, we generate one story based on its definition with each LLM. We recruit human subjects with law education backgrounds on Prolific to evaluate the legal concepts and their corresponding stories generated by LLMs. Due to budget constraints, we randomly sampled 20 out of 101 concepts to compare among LLaMA 2, GPT-3.5, and GPT-4. Subsequently, we compare their performance and evaluate the full set of 101 concepts on the best model. For human evaluation, we recruit three raters to judge the \textbf{Readability of Definition (RoD)} and the following metrics for the generated stories: (1) \textbf{Readability of Story (RoS)}, (2) \textbf{Relevance}, (3) \textbf{Redundancy}, (4) \textbf{Cohesiveness}, (5) \textbf{Completeness}, (6) \textbf{Factuality}, (7) \textbf{Likeability}, (8) \textbf{Believability}. We use a 5-item Likert scale where 1 means very bad and 5 means very good. Details about the metrics and human evaluation are discussed in Appendix \ref{appendix:human_prolific}.

\paragraph{Results} In Table \ref{tab:story_human_eval}, we have several interesting observations. First, GPT-4 outperforms GPT-3.5 and LLaMA 2 in almost all the metrics except for redundancy. LLaMA 2 performs slightly better than GPT-3.5 in most metrics. By examining stories generated by LLaMA 2, we find that 8 (out of 20) generated stories are not in a story style but simplified definitions in plain language. Therefore, ``stories'' generated by LLaMA 2 seem shorter and more concise with high redundancy scores. In contrast, all the ``stories'' generated by GPT-3.5 and GPT-4 are indeed stories. 
Second, we observe consistently higher readability scores in stories (RoS) than the definitions (RoD), indicating that people find these stories easier to read than legal language. Additionally, GPT-4 stories receive high scores ($\geq 4.5$) in both readability and cohesiveness, showing they are easy to read in story format. Third, GPT-4 stories also achieve high scores ($\geq 4.5$) in relevance, completeness, and factuality, meaning that these stories are relevant to the definitions, and have good coverage and faithfulness reflection in the definition. Finally, human annotators find these stories decently likable ($\mu = 4.42$) and believable ($\mu = 4.48$). In practice, the stories can be further refined through expert feedback if their quality is not good enough. However, given the high ratings these generated stories have received, we have decided to provide them to participants along with their respective definitions, without the need for additional expert reviews.



\subsubsection{Complexity Evaluation}
Legalese usually contains long, wordy, complicated sentence structures, making it difficult for the public to understand. The readability metric in the previous section is one way to assess this. We also use multiple automatic measures of language complexity to compare concept definitions from Wikipedia and stories generated by different LLMs. These measures are not meant to be exhaustive but to provide more nuanced insights into language complexity, which is important for reader comprehension. We report the following common complexity metrics for comparison: (1) \textbf{Legal Vocabulary List (LVL) occurrences}, (2) \textbf{Top 1000 most common words (Top1K)}, (3) \textbf{Function words}, (4) \textbf{Sentence length}, (5) \textbf{Language model perplexity}, (6) \textbf{Flesch-Kincaid grade level}. More metric details can be found in Appendix \ref{appendix:complexity}.

\paragraph{Results} In Table \ref{tab:complexity_eval}, we compare the linguistic complexity between Wikipedia definitions and LLM-generated stories. We observe that stories from GPT-4 contain the lowest LVL proportion, GPT perplexity, sentence length, and Flesch-Kincaid score. Both GPT-4 and GPT-3.5 tend to use more function words and the top 1000 words from Thing Explainer in the stories. Across different measures, the definitions have the most linguistic complexity. Stories from GPT-3.5 and GPT-4 use a language with similar complexity but simpler than those from LLaMA 2. These observations are consistent with the RoD and RoS scores from the human evaluation in Table \ref{tab:story_human_eval}.

\setlength{\extrarowheight}{2pt}
\begin{table}[t!]
    \centering
    \footnotesize
    \resizebox{0.45\textwidth}{!}{%
    \begin{tabular}{ccccc}
        {\textbf{Metrics}}  & 
        {\textbf{Wiki}}  &
        {\textbf{LLaMA 2}}  &
        {\textbf{GPT-3.5}} & 
        {\textbf{GPT-4}} \\
        \Xhline{3\arrayrulewidth}
        LVL & $0.27$ & $0.26$ & $0.22$ & $\mathbf{0.21}$ \\
        Top1K & $0.51$ & $0.61$ & $\mathbf{0.64}$ & $0.63$ \\
        Func. Words & $0.37$ & $0.40$ & $\mathbf{0.42}$ & $0.41$ \\
        GPT PPL. & $65.82$ & $30.51$ & $25.20$ & $\mathbf{24.96}$ \\
        Sent. Length & $30.11$ & $23.41$ & $20.26$ & $\mathbf{19.61}$ \\
        FK Scores & $14.79$ & $11.35$ & $8.61$ & $\mathbf{8.23}$ \\

    \end{tabular}
    }
    \caption{Results in complexity metrics in Wikipedia definitions and LLM-generated stories on the same subset of 20 concepts. Wiki represents the definition from Wikipedia. LLaMA 2, GPT-3.5 and GPT-4 represent stories generated by these models. The numbers in bold suggest the highest readability in that particular metric.}
    \label{tab:complexity_eval}
\end{table}

\subsection{Question Evaluation}
It is challenging to evaluate generated questions since there are no gold standard questions. Therefore, we use human evaluation to assess the quality of generated questions and rely on the critique of two legal experts for improvement. We first begin with three authors of this work examining a subset of 10 concepts (30 questions), documenting emerging noticeable errors, and, following rounds of discussion, summarizing these errors as a set of six common error categories. These error categories and an ``other'' option for non-categorized errors are used to facilitate the evaluation process with Prolific workers and two legal experts. 

\subsubsection{Human Evaluation}
Specifically, we recruit three Prolific human evaluators with law knowledge to judge whether the question contains the following shortcomings: (1) the question is too easy and simple, (2) the answer cannot be derived from the definition or story above, (3) the question is confusing, (4) there are more than one right answer in the 4 options, (5) there is no right answer among the 4 options, (6) the reasoning given in the suggested answer is wrong or flawed, (7) other issues not covered above. The annotators can select more than one option if multiple issues are identified. If there is no error, they can choose (8) There is no issue. The annotators also rate each question from 1 (bad) to 5 (good).

\paragraph{Human Ratings}
In Table \ref{tab:question_human_eval}, we find GPT-4 generation outperforms the other LLMs across all three types of questions. LLaMA 2 performs slightly better than GPT-3.5. In addition, the results show that concept questions and prediction questions have much higher scores than the limitation questions. We hypothesize that limitation questions are challenging for LLMs to generate because not all concepts are apparent to discuss their limitations or exceptions. 
The disparities in the question quality and error rate necessitate human-in-the-loop methods to improve question generation and quality control. 

\setlength{\extrarowheight}{2pt}
\begin{table}[t]
    \centering
    \footnotesize
    \resizebox{0.45\textwidth}{!}{%
    \begin{tabular}{cccc}
        {\textbf{Model}}  & 
        {\textbf{ConceptQ}}  &
        {\textbf{PredictionQ}} & 
        {\textbf{LimitationQ}} \\
         \Xhline{3\arrayrulewidth}
        \multicolumn{4}{c}{\textit{101 Concepts (Mean\textsubscript{STD})}} \\ 
        GPT-4 & $4.46_{0.77}$ & $4.35_{0.78}$ & $4.14_{0.98}$ \\
        \Xhline{3\arrayrulewidth}
        \multicolumn{4}{c}{\textit{Sampled 20 Concepts (Mean\textsubscript{STD})}} \\ 
        GPT-4 & $\mathbf{4.47}_{0.70}$ & $\mathbf{4.35}_{0.71}$ & $\mathbf{4.27}_{0.84}$ \\
        GPT-3.5 & $4.12_{0.64}$ & $3.95_{0.85}$ & $3.48_{0.91}$ \\
        LLaMA 2 & $4.23_{0.96}$ & $4.10_{1.22}$ & $4.12_{1.12}$ \\
    \end{tabular}
    }
    \caption{Human evaluation on LLM-generated educational questions. Three questions are generated per concept, including a concept question, a prediction question, and a limitation question. The upper section contains scores for GPT-4 on the complete 101 legal concepts. The lower section contains scores for GPT-4, GPT-3.5, and LLaMA 2 on a subset of 20 legal concepts.}
    \label{tab:question_human_eval}
\end{table}

\begin{figure}[t]
\centering
\includegraphics[width=\linewidth]{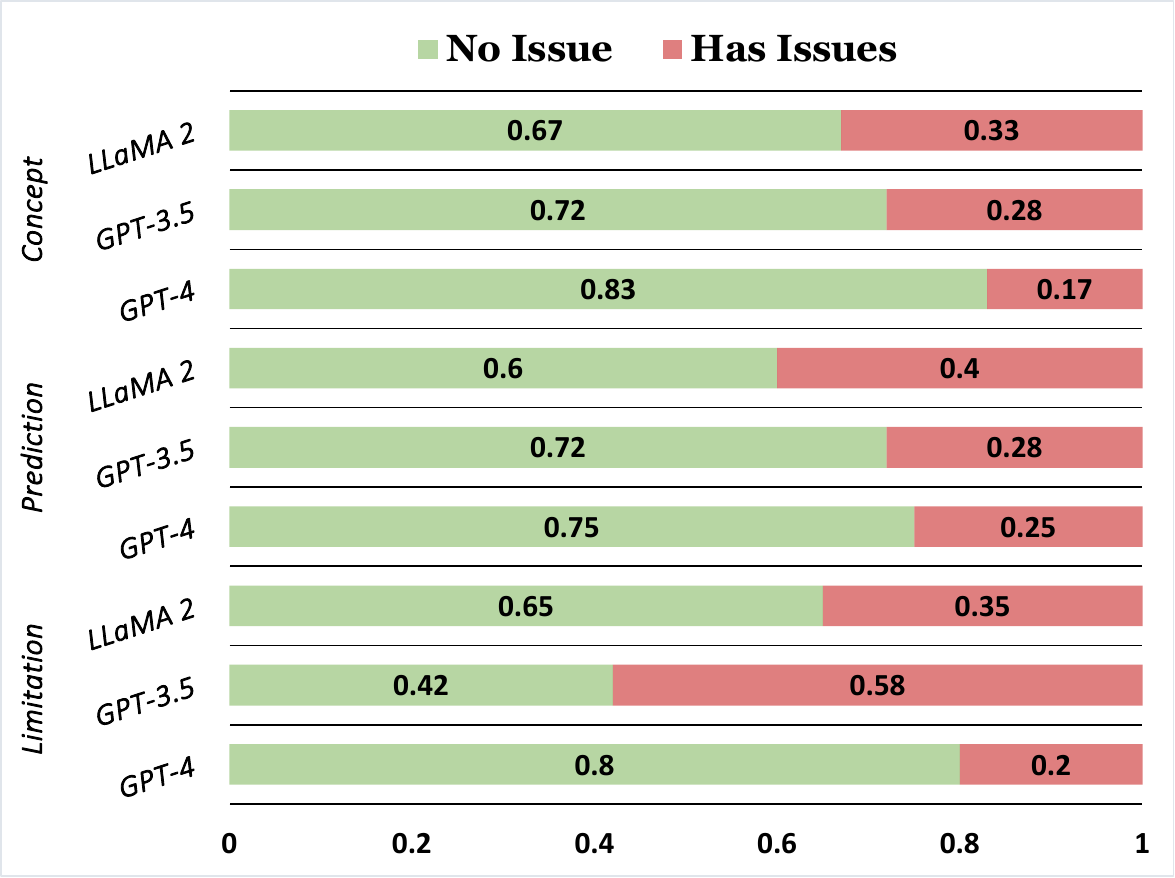}
  \captionof{figure}{Distribution of questions with or without issues generated by LLaMA 2, GPT-3.5, and GPT-4.}
  \label{fig:no_issue_distri}
\end{figure}

\paragraph{Error Analysis}
In Figure \ref{fig:no_issue_distri}, with the human evaluation of LLMs-generated questions for 20 sampled concepts, we compute the percentage of questions with issues versus those without any. GPT-4 outperforms other LLMs in generation questions with no issues at 83\%, 75\%, and 80\% for concept, prediction, and limitation questions, respectively.

\begin{figure}[t]
\centering
\includegraphics[width=\linewidth]{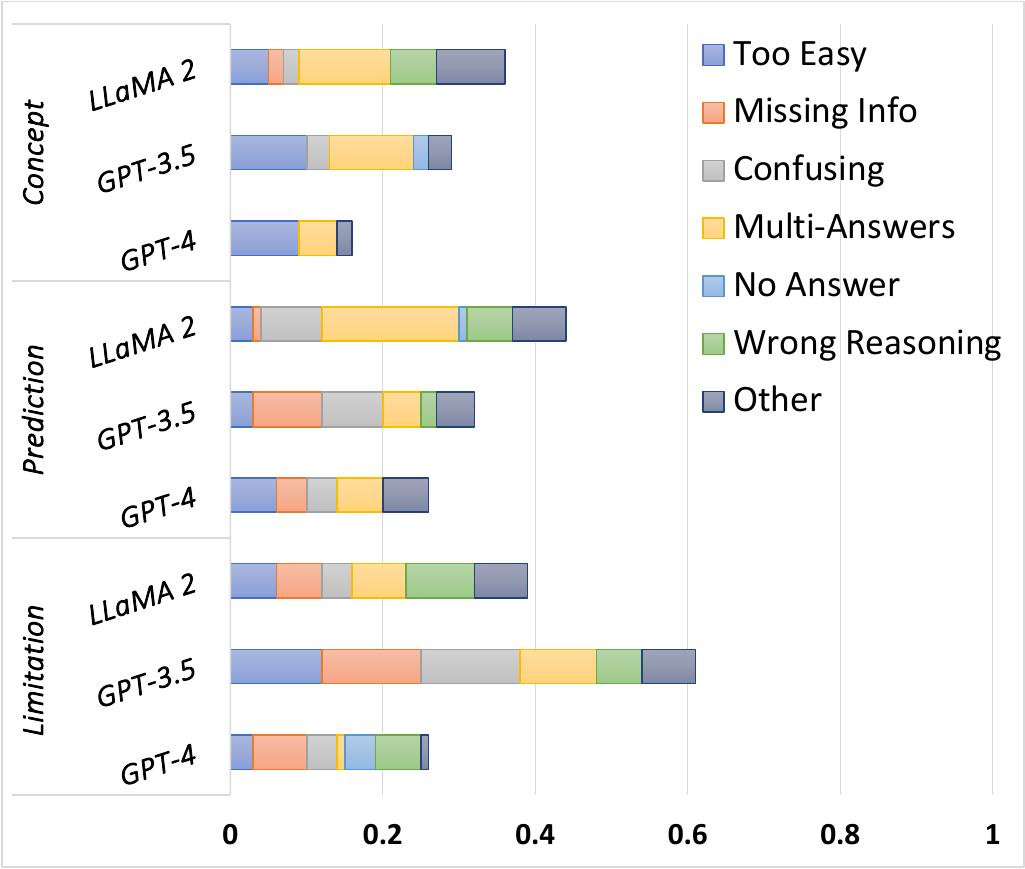}
  \captionof{figure}{Distribution of different issues among the questions generated by LLaMA 2, GPT-3.5, and GPT-4.}
  \label{fig:error_distri}
\end{figure}

Furthermore, we break down the distribution of errors found by the human annotators in the multiple-choice questions in Figure \ref{fig:error_distri}. We account for all the issue labels as we have allowed the annotators to select more than one issue. We observe that the six labels provided to the annotators cover most of the errors in the questions (the option ``Other'' is chosen when the labels do not cover certain errors). As shown in Figure \ref{fig:error_distri}, each LLM makes a combination of different errors; for instance, GPT-4 created fewer confusing questions and less wrong/flawed reasoning for the correct answer than other LLMs.
\subsubsection{Expert Critiques}
\vspace{-1mm}
Finally, we recruit two legal experts to evaluate the efficacy of the assessment with critiques by completing the assessment on the 20 sampled concepts. The purpose of this step is to make sure the questions are answerable, given either definitions alone or definitions and stories. We randomly split 20 sampled concepts into two equal batches to avoid exposure bias. Each person completes one batch with the definition only and the other batch with the story and definition. However, the batches for two people are in reversed conditions. With this design, we can compare their agreeability without the exposure bias from the stories. At last, they are asked to categorize the difficulty level of each legal concept into easy, medium, and hard. 

\paragraph{Inter-rater Agreeability} We observe that answers from the two legal experts have overall high agreement scores in Cohen's Kappa with GPT-4 answers (ranging from 0.77 to 1.00) and each other (ranging from 0.68 to 0.86) across different question types. We include agreement scores in Table \ref{tab:expert_question_eval}. The concept question has the highest agreement scores. This confirms our finding from the previous human ratings from Prolific annotators in Table \ref{tab:question_human_eval}, which shows that the concept question has the highest average human ratings. Expert 1 gave improvement advice to 6 questions and Expert 2 to 2 questions. After we prompt GPT-4 with their advice to re-generate the corresponding questions, we show them to the experts, and they are 100\% approved by the experts after the first round of regeneration. This shows the effectiveness of instructing GPT-4 with expert advice to revise the questions (see one example in Appendix \ref{appendix:expert_generation_sample}).




\section{RCT Experiment}
\subsection{Experiment Design}
Using the stories and questions from the pipeline, we evaluate the efficacy of LLM-generated stories for helping human comprehension on a sample of 10 concepts. We design a randomized controlled trial (RCT) with two study conditions: (1) a control group that is given the legal concept definition and (2) a treatment group that is given both the definition and a story that illustrates a hypothetical situation in which the given concept applies. Both condition groups read the content and are asked to complete three multiple-choice questions to evaluate comprehension. We keep both definition and story in the treatment group (instead of just the story) because we hope to have the definition as a good reference if the story is confusing to the reader by any chance. Due to the importance of legal education for both native and non-native English speakers (such as immigrants), we recruit participants from both native-speaker and non-native speaker populations. The participants were recruited through Prolific; details of the criteria are in Appendix \ref{appendix:rct_details}. We recruit 15 to 20 people in each condition to complete batches of 5 concepts. In total, we have 65 respondents from native speakers ($33$ in control and $32$ in treatment) and 71 respondents from non-native speakers ($37$ in control and $34$ in treatment conditions) for a total of 136 participants. With these participants, we collect the following information through 5-point Likert scales: (1) the relevance of the legal concept, (2) their interest in engaging and learning more about the concept, (3) their familiarity with the concept, (4) their familiarity with the scenario setting in the story (treatment group only). Prior studies \cite{glonek2014listening,kromka2019classroom} show that stories facilitate learner recall about content. The participant comprehension is evaluated by their accuracy in RC questions and learning retention in the three-day post-study assessment, in which we present the participants with the same RC questions but with the answers reordered, without providing them with the definition or story again.

\subsection{Results \& Discussions}

\paragraph{Reading Comprehension} 
We compare the answer accuracy in Table \ref{tab:rct_accuracy} and find that \textbf{legal stories improve the comprehension accuracy for the non-native English speaker group for all question types while only improving the limitation question accuracy for the native speaker group.} Specifically, the concept and prediction accuracy for native speakers decrease in the treatment group. However, the Chi-Squared test for statistical significance fails to show significant differences between two conditions in native English speakers (concept question: $X^2 (1, N = 325) = 0.48, p = .487$; prediction question: $X^2 (1, N = 325) = 0.65, p = .419$; limitation question: $X^2 (1, N = 325) = 2.01, p = .156$). For the non-native speaker group, statistical significance is achieved for both prediction and limitation questions (concept question: $X^2 (1, N = 355) = 0.20, p = .653$; prediction question: $X^2 (1, N = 355) = 4.29, \textbf{p = .038}$; limitation question: $X^2 (1, N = 355) = 11.77, \textbf{p < .001}$). We have summarized individual-level mean and standard deviation in accuracy in Appendix \ref{appendix:individual_accuracy}.

The difference in accuracy for the native speakers between the treatment and control conditions is not statistically significant; thus, it could be due to the sampling error. We also notice that the native speakers' accuracy for the concept and prediction questions in the control group is the highest or close to the highest accuracy among other groups, which might imply a ceiling effect of the learning interventions. The stories may have enhanced the participant's accuracy in questions with originally lower accuracy in the control setting. For instance, both native and non-native speakers had more room for improvement in the control condition, which the stories could have potentially addressed.

\setlength{\extrarowheight}{2pt}
\begin{table}[t]
    \centering
    \footnotesize
    \resizebox{0.45\textwidth}{!}{%
    \begin{tabular}{cccc}
        {\textbf{Condition}}  & 
        {\textbf{ConceptQ}}  &
        {\textbf{PredictionQ}} & 
        {\textbf{LimitationQ}} \\
         \Xhline{3\arrayrulewidth}
        \multicolumn{4}{c}{\textit{Native Speakers (Accuracy)}} \\ 
        Definition & $\mathbf{93.33}$ & $\mathbf{78.79}$ & $77.58$ \\
        Def. + Story & $90.62$ & 74.38 & $\mathbf{84.38}$ \\
        \Xhline{3\arrayrulewidth}
        \multicolumn{4}{c}{\textit{Non-native Speakers (Accuracy)}} \\ 
        Definition & $89.19$ & $71.89$ & $68.65$ \\
        Def. + Story & $\mathbf{91.18}$ & $\mathbf{81.76}$ & $\mathbf{84.71}$ \\
    \end{tabular}
    }
    \caption{Comprehension accuracy of native and non-native speakers of English. We report their performance in concept, prediction, and limitation questions.}
    \label{tab:rct_accuracy}
\end{table}

\setlength{\extrarowheight}{2pt}
\begin{table}[t]
    \centering
    \footnotesize
    \resizebox{0.45\textwidth}{!}{%
    \begin{tabular}{cccc}
        {\textbf{Condition}}  & 
        {\textbf{ConceptQ}}  &
        {\textbf{PredictionQ}} & 
        {\textbf{LimitationQ}} \\
         \Xhline{3\arrayrulewidth}
        \multicolumn{4}{c}{\textit{Native Speakers (Retention Rate)}} \\ 
        Definition & $\mathbf{92.55}$ & $\mathbf{88.89}$ & $\mathbf{91.03}$ \\
        Def. + Story & $91.58$ & $86.84$ & $91.01$ \\
        \Xhline{3\arrayrulewidth}
        \multicolumn{4}{c}{\textit{Non-native Speakers (Retention Rate)}} \\ 
        Definition & $86.32$ & $82.80$ & $91.01$ \\
        Def. + Story & $\mathbf{98.56}$ & $\mathbf{89.60}$ & $\mathbf{92.31}$ \\
    \end{tabular}
    }
    \caption{Retention rate in delayed post-study assessment of native and non-native speakers of English.
    The highest possible percentage is 100\%, representing a perfect retention of knowledge (in theory).}
    \label{tab:retention_rate}
\end{table}

\paragraph{Relevance} The relevance score, which rates the degree to which the content displayed to participants is relevant to their own lives and situations, shows that \textbf{participants who read the stories with the concept definition consistently felt it more relatable than the control condition for both native speaker and non-native speaker groups}. Among non-native English speakers, we observe that participants who read both definitions and stories find these legal concepts more relevant ($3.19_{1.17}$ vs. $2.47_{1.17}$) to their lives than those who read only definitions. Similarly, we observe higher relevance scores ($3.21_{1.32}$ vs. $2.63_{1.30}$) for the treatment group compared with the control group among native speakers. Mann-Whitney U tests find statistical significance in relevance scores for both native English speakers ($U = 16421.0, \textbf{p < .001}$) and non-native English speakers ($U = 21077.5, \textbf{p < .001}$).

\paragraph{Interest in Law} The interest score, which rates the degree to which participants feel interested in engaging with the content displayed, shows that \textbf{the treatment condition has statistically higher interests than the control condition for the non-native speaker group but not for the native-speaker group.}  Specifically, among non-native English speakers, we observe that participants who read both definitions and stories are more interested in delving into laws and legal knowledge ($4.03_{1.20}$ vs. $3.84_{0.92}$) than those who read only definitions. Among native English speakers, the treatment group shows slightly higher interest than the control group ($3.78_{0.99}$ vs. $3.67_{1.12}$). Mann-Whitney U tests reveal that statistical differences are found for the participant interest between the treatment group and control group for non-native speakers ($U = 18662.5, \textbf{p = .001}$); however, not for the native English speaker ($U = 13800.0, p = .459$).

\paragraph{Knowledge Retention} To investigate participants' comprehension retention, we send the follow-up assessment three days after the original study. In the end, 71\% of the respondents filled out the delayed post-study test. Given participants' original and delayed assessment, the retention rate is calculated as the percentage of continued correct answers for each question. The result is summarized in Table \ref{tab:retention_rate}. We observe that after three days, participants show different degrees of forgetting. However, \textbf{non-native speakers who read stories and definitions have a higher retention rate after three days while no such effect is found for the native speaker group}. A Chi-square test confirms significant differences for non-native speakers between the retention in the treatment and control group for the concept question $X^2 (1, N = 256) = 12.74, \textbf{p < .001}$; however, no such differences are found for the other questions.

\section{Conclusion}
In this work, we explore a novel application of LLMs in legal concept learning through storytelling. We use an expert-in-the-loop pipeline to create the \textsc{LegalStories} dataset that contains educational stories and comprehension questions. Moreover, we compare the performance of several benchmark LLMs including GPT-4, GPT-3.5, and LLaMA 2 with automatic and human evaluations. While GPT-4 outperforms the others at generating legal stories and creating questions, it still exhibits certain reasoning errors, highlighting the need for human supervision when using LLMs for educational content development. Finally, through RCTs, we show that, among non-native speakers, learning with stories not only improves comprehension of legal concepts and interest but also leads to a higher retention rate in the follow-up assessment compared to learning with definitions alone.
Our study suggests considerable potential for using LLMs in advancing legal education and beyond.

\section*{Limitations}




\paragraph{Sample Size} Given the limited financial budget available to conduct our research, we chose to conduct our study in a smaller data setting to obtain high-quality human feedback. Similarly, our participant pool was limited by extensive and costly surveying approaches to 65 native respondents and 71 non-native respondents, which may have negatively impacted the statistical power of our group comparison results. We want to emphasize that even at this scale the cost is nontrivial. For example, it took around 800 dollars to evaluate 101 concepts generated by GPT-4 and 320 dollars to evaluate 20 concepts generated by GPT-3.5 and LLaMA-2 in total. For human experiments, it took around 2400 dollars to run RCTs including the follow-up study with 10 sampled concepts. In an attempt to alleviate this, we have carefully chosen our current sampling strategy to ensure that the resulting samples are representative. For instance, we compare the human evaluation between 101 and 20 concepts in Table \ref{tab:story_human_eval} and Table \ref{tab:question_human_eval} and show that they are similar for GPT-4. Although similar studies exist with similar group sizes per condition \cite{lu2021expert,steuer2022educational,august-etal-2022-generating}, and although we find our results compelling, it is possible that due to sample size limitations, we were not able to capture small effects; however, our study provides strong statistical power for observing large effect sizes, revealing several significant effects, even with limited statistical power.

\paragraph{Data Quality \& Practicality} To mitigate biases and hallucinations of LLM-generated content, we loop in crowdworkers and experts to audit and improve the generated content. Legal experts estimate that it takes at least 30 minutes to create one story and three questions from scratch for each concept. By contrast, it takes each expert around 6.5 minutes to evaluate and write feedback to each legal concept, story, and questions. Although it is a labor-saver compared to having a completely human-written dataset, it still requires human experts, thus, might suffer from scalability. In practice, we believe that having human experts (such as teachers and lawyers) in the loop is a reliable and necessary manner to create useful and less biased educational content by mitigating model errors while minimizing human effort. 

\paragraph{RCT Design} In our case, due to limited resources, we were unable to run multiple human evaluations to optimize our prompts. We also chose the most intuitive control group (definition) and treatment group (definition+story) to answer the main research question whether LLM-generated stories improve comprehension in legal concepts. The control group represents current practice for legal communication (e.g., a glossary of terms or a legal dictionary). We believe that it would be interesting for future studies to extend the work by comparing this with various prompting strategies to generate LLM-based concept explanations or elaborations such as ``Explain Like I'm 5 (ELI5)''.

\section*{Ethical Considerations}

\paragraph{Legal Experts} The term ``legal expert'' does not constitute any suggestion or indication that the participants enlisted to provide evaluation and critique of the LLM-generated materials are admitted to practice law in any jurisdictions or are holding themselves out as attorneys qualified to provide legal advice. Their participation in this research does not involve legal representation, legal advice, or drafting of legal documents for any entity or person to any extent. The legal experts enlisted either have graduated with law degrees or have made substantial progress toward earning law degrees. As such, the participants can provide valuable feedback for the research because their legal training makes them better-suited to assess LLM-generated content than people without such backgrounds.

\paragraph{Code of Conduct} This research follows the ACL Code of Ethics, has IRB Exempt status, and respects participants' anonymity. We used the Prolific platform for human annotation and experiments with their consent, compensated online annotators $\$$15 per hour according to Massachusetts state law, and ensured LLM-generated content is safe and non-offensive. Exact experiment details are included in the appendix for reproducibility.

\paragraph{LLM-related Risks} We are aware of the potential for bias that LLMs present, both in educational and in generalized contexts. It is dangerous and inappropriate to provide LLM contents to students without human supervision, because these contents might contain misleading, biased, harmful, or wrong information and education is a high-stake domain. With respect to risks such as this, our work takes a human-centered approach to loop in qualified crowdworkers and experts to audit the LLM outputs. In practice, we believe that having human experts such as teachers and lawyers in the loop is a reliable and effective manner to create useful and less biased high-quality educational content. 




\paragraph{Information Loss} Law is, by nature, a sensitive domain, and computational tools must be designed responsibly. In the context of legal education, it is critical to design comprehension tools in ways that do not over-simplify or over-generalize the nuances of legal jargon. To address these issues, we chose to collaborate with legal experts to audit the content and draw on domain-specific data, we hope to provide an approach that balances access to justice needs with responsible AI approaches.


\section*{Acknoledgement}
We would like to thank Maria Antoniak, Bailin Wang, Belén Saldías, and Mingye Gao for their helpful discussions. We also thank the reviewers from ACL Rolling Review (ARR) for their constructive feedback.

\bibliography{anthology,custom}

\appendix
\onecolumn

\section{Dataset Generation}
\subsection{Model Details for Story and Question Generation}
\label{appendix:models}
LLaMA 2 (\texttt{LLaMA 2-70b-Chat}), GPT-3.5 (\texttt{GPT-3.5-turbo-0613}), and GPT-4 (\texttt{GPT-4-0613}) are used for this experiment. For LLaMA 2, we set top p to 1.0 and temperature to 0.01 and use the default settings for the other parameters. For GPT-3.5 and GPT-4, we set the temperature to 0.0 and use the default settings for the other parameters. We use the Replicate LLaMA 2 API \footnote{\url{https://replicate.com/meta/llama-2-70b-chat}} in our experiments. The exact prompt for the story generation is shared in the main paper. We also include the exact prompts for question generation in the subsection below. 

\subsection{Prompt for Story Generation}
\label{appendix:prompt_story}
Here is the exact prompt used to generate stories, where \{CONCEPT\} stands for the concept name and \{DEFINITION\} for the Wikipedia definition: Tell a story within 500 words to simplify the concept explanation below for ``\{CONCEPT\}''. Start your answer with ``Concept Simplified:''. Concept: ``\{DEFINITION\}''.

\subsection{Prompts for Question Generation}
\label{appendix:prompts}
We present the prompts used to generate each multiple-choice question and answers in the question generation phase of the pipeline. \{CONCEPT\} stands for the concept name, \{DEFINITION\} for the Wikipedia definition, and \{STORY\} for the generated story from the corresponding LLM. 

\paragraph{Concept Question Prompt} Read the concept explanation and story below. Please generate a multiple choice question with four candidates (only one correct answer) to test if a reader understands the concept: \textbf{which of the following answers is an accurate description of the concept ``\{CONCEPT\}''}? Start your response with ``Question:''. Candidates are ordered by (A), (B), (C), (D). In the end, give the right answer with its explanation starting with ``The right answer is''. Concept: ``\{DEFINITION\}''. Story: ``\{STORY\}''

\paragraph{Prediction Question Prompt} Read the concept explanation and story below. Please generate a multiple choice question with four candidates (only one correct answer) to test if a reader understands the concept: \textbf{come up with a hypothetical scenario where the concept ``\{CONCEPT\}'' is used and ask the reader to guess the ending of the story}. Please ensure the hypothetical scenario is more challenging than the story below. Start your response with ``Question:''. Candidates are ordered by (A), (B), (C), (D). In the end, give the right answer with its explanation starting with ``The right answer is''. Concept: ``\{DEFINITION\}''. Story: ``\{STORY\}''

\paragraph{Limitation Question Prompt} Read the concept explanation and story below. Please generate a multiple choice question with four candidates (only one correct answer) to test if a reader understands the concept: \textbf{what is a potential limitation or exception of the rule ``\{CONCEPT\}''}? Start your response with ``Question:''. Candidates are ordered by (A), (B), (C), (D). In the end, give the right answer with its explanation starting with ``The right answer is''. Concept: ``\{DEFINITION\}''. Story: ``\{STORY\}''

\subsection{Expert-LLM Agreement}

Table \ref{tab:expert_question_eval} shows the Cohen's Kappa agreement scores between two legal experts and GPT-4 in choosing the right answers in generated questions. 

\setlength{\extrarowheight}{2pt}
\begin{table}[ht!]
    \centering
    \footnotesize
    \begin{tabular}{ccc}
        {\textbf{Expert 1 \& GPT-4}}  &
        {\textbf{Expert 2 \& GPT-4}} & 
        {\textbf{Expert 1 \& Expert 2}} \\
         \Xhline{3\arrayrulewidth}
        \Xhline{1\arrayrulewidth}
        \multicolumn{3}{c}{\textit{Concept Question (Cohen's Kappa)}} \\ 
        $1.00$ & $0.86$ & $0.86$ \\

        \Xhline{2\arrayrulewidth}
        \multicolumn{3}{c}{\textit{Prediction Question (Cohen's Kappa)}} \\ 
        $0.92$ & $0.77$ & $0.68$ \\

        \Xhline{2\arrayrulewidth}
        \multicolumn{3}{c}{\textit{Limitation Question (Cohen's Kappa)}} \\ 
        $0.85$ & $0.78$ & $0.86$ \\
    \end{tabular}
    \caption{We compare expert and GPT-4 answers on LLM-generated educational questions of 20 legal concepts. We show Cohen's Kappa scores among two legal experts and GPT-4.}
    \label{tab:expert_question_eval}
\end{table}

\subsection{\textsc{LegalStories} Example with Expert Feedback}
\label{appendix:expert_generation_sample}

\subsubsection{Story Example}
We show one example concept ``Substantial Certainty Doctrine'' from the \textsc{LegalStories} dataset:

\paragraph{Doctrine:} \textit{Substantial Certainty Doctrine}
\paragraph{WiKi Definition:} In law, the substantial certainty doctrine is the assumption of intent even if the actor did not intend the result, but knew with substantial certainty the effect would occur as a result of his action. The doctrine can be used by courts as a test to determine whether or not a defendant committed a tort. For example, in Garratt v. Dailey (1955), the Washington Supreme Court remanded a case back to the lower courts to determine whether or not the five year-old defendant ``knew with substantial certainty that the plaintiff would attempt to sit down where the chair which he moved had been.''
\paragraph{Story:} Imagine you're at a party and you see your friend about to sit on a chair. You quickly pull the chair away just as they're about to sit, resulting in them falling on the floor. You didn't intend for them to fall and get hurt, but you knew there was a high chance it would happen because you pulled the chair away. This is the idea behind the ``substantial certainty doctrine'' in law.

This legal concept suggests that even if you didn't mean for a certain outcome to happen, if you knew there was a high chance it would, you can still be held responsible. Courts use this doctrine to decide if someone committed a wrong act, also known as a tort.

A famous example of this is the case of Garratt v. Dailey in 1955. In this case, a five-year-old boy moved a chair just as an elderly woman was about to sit. The woman fell and got injured. The court had to decide if the boy ``knew with substantial certainty'' that the woman would try to sit where the chair had been and get hurt. The case was sent back to the lower courts to make this decision.

So, in simple terms, the substantial certainty doctrine is like saying, ``You knew what would happen if you did that, so you're responsible for the outcome.''
\paragraph{Concept Question:} Which of the following best describes the ``Substantial Certainty Doctrine'' in law?
\begin{enumerate}[label=(\Alph*)]
    \item It is a doctrine that holds a person responsible for an outcome only if they intended for that outcome to occur.
    \item It is a doctrine that holds a person responsible for an outcome if they knew there was a high chance it would occur, even if they did not intend for it to happen.
    \item It is a doctrine that holds a person responsible for an outcome only if they did not know there was a high chance it would occur.
    \item It is a doctrine that holds a person responsible for an outcome only if they did not intend for that outcome to occur, and they did not know there was a high chance it would happen.
\end{enumerate}

The right answer is \textbf{(B)}. The ``Substantial Certainty Doctrine'' in law holds a person responsible for an outcome if they knew there was a high chance it would occur, even if they did not intend for it to happen. This doctrine is used by courts to determine if someone committed a tort, or a wrongful act.

\paragraph{Prediction Question:} In a hypothetical scenario, John, a construction worker, removes a safety barrier around a deep pit at a construction site to make his work easier. He sees his co-worker, Mike, walking towards the pit while talking on the phone, unaware of the danger. John doesn't warn Mike about the pit, thinking Mike would notice it himself. However, Mike falls into the pit and gets severely injured. If this case goes to court, under what condition might the substantial certainty doctrine be applied?
\begin{enumerate}[label=(\Alph*)]
    \item The court would apply the substantial certainty doctrine if it can be proven that John knew with substantial certainty that Mike would fall into the pit.
    \item The court would apply the substantial certainty doctrine if it can be proven that John intentionally removed the safety barrier.
    \item The court would not apply the substantial certainty doctrine because John did not intend for Mike to fall into the pit.
    \item The court would not apply the substantial certainty doctrine because Mike should have been more careful.
\end{enumerate}
The right answer is \textbf{(A)}. The substantial certainty doctrine could be applied if it can be proven that John knew with substantial certainty that Mike would fall into the pit. Even though John did not intend for Mike to fall, if he was aware of the high risk of such an outcome when he removed the safety barrier and did not warn Mike, this aligns with the substantial certainty doctrine, which assumes intent even if the actor did not intend the result, but knew with substantial certainty the effect would occur as a result of his action.

\paragraph{Limitation Question:} What is a potential limitation or exception of the substantial certainty doctrine?
\begin{enumerate}[label=(\Alph*)]
    \item The doctrine cannot be applied if the defendant is a minor.
    \item The doctrine cannot be applied if the defendant did not physically cause the harm.
    \item The doctrine cannot be applied if the defendant did not have a direct intention to cause harm.
    \item The doctrine cannot be applied if the defendant did not know with substantial certainty that their action would cause the effect.
\end{enumerate}

The right answer is \textbf{(D)}. The doctrine cannot be applied if the defendant did not know with substantial certainty that their action would cause the effect. The substantial certainty doctrine assumes intent even if the actor did not intend the result, but knew with substantial certainty the effect would occur as a result of his action. If the defendant did not have this level of certainty, the doctrine may not apply.

\subsubsection{Question before and after Expert Feedback}
\label{appendix:expert_feedback}
In this section, we walk through one example question to explain how expert feedback is used for re-generating multiple-choice questions with quality control. The prompt we used is as follows: ``\{EXPERTISE\_SUGGESTION\}. Improve the question given the feedback above.'' The example we use is the \textbf{Prediction Question} for \textit{Substantial Certainty Doctrine}. The original prediction question generated by LLM is as follow:
\paragraph{Original Generation:} 
In a hypothetical scenario, John, a construction worker, removes a safety barrier around a deep pit at a construction site to make his work easier. He sees his co-worker, Mike, walking towards the pit while talking on the phone, unaware of the danger. John doesn't warn Mike about the pit, thinking Mike would notice it himself. However, Mike falls into the pit and gets severely injured. If this case goes to court, how might the substantial certainty doctrine be applied?
\begin{enumerate}[label=(\Alph*)]
    \item The court would not apply the substantial certainty doctrine because John did not intend for Mike to fall into the pit.
    \item The court would apply the substantial certainty doctrine because John knew with substantial certainty that Mike could fall into the pit.
    \item The court would not apply the substantial certainty doctrine because Mike should have been more careful.
    \item The court would apply the substantial certainty doctrine because John intentionally removed the safety barrier.
\end{enumerate}
To this question, the answer generated by LLM is:
\begin{quote}
    The right answer is (B). The substantial certainty doctrine could be applied because John knew with substantial certainty that Mike could fall into the pit. Even though John did not intend for Mike to fall, he was aware of the high risk of such an outcome when he removed the safety barrier and did not warn Mike. This aligns with the substantial certainty doctrine, which assumes intent even if the actor did not intend the result, but knew with substantial certainty the effect would occur as a result of his action.
\end{quote}

\paragraph{Expert Critiques} The legal expert reviewed the multi-choice question and gave the following feedback:
\begin{enumerate}
    \item Does your answer agree with the suggested answer?
    
    \textit{Response}: No.
    \item Is the suggested answer correct or wrong? Why? 

    \textit{Response}: The suggested answer seems wrong.
    \item What suggestions do you have to improve the question?

    \textit{Response}: I think the suggested answer is wrong/the question is a bit confusing to me. When John removes the barrier, he does not know with substantial certainty that Mike would fall in.
\end{enumerate}

From this expert critique, we insert the following prompt to LLM for question re-generation: 
\begin{quote}
    \textit{I think the suggested answer is wrong/the question is a bit confusing to me. When John removes the barrier, he does not know with substantial certainty that Mike would fall in.}
    
    \textit{Improve the question given the feedback above.}
\end{quote}

Afterwards, the LLM takes into the critiques in the prompt, fixes the confusion about whether Jon knows the substantial certainty of his action to Mike, and produces new choices without changing the question itself:
\paragraph{Final Prediction Question:} In a hypothetical scenario, John, a construction worker, removes a safety barrier around a deep pit at a construction site to make his work easier. He sees his co-worker, Mike, walking towards the pit while talking on the phone, unaware of the danger. John doesn't warn Mike about the pit, thinking Mike would notice it himself. However, Mike falls into the pit and gets severely injured. If this case goes to court, under what condition might the substantial certainty doctrine be applied?
\begin{enumerate}[label=(\Alph*)]
    \item The court would apply the substantial certainty doctrine if it can be proven that John knew with substantial certainty that Mike would fall into the pit.
    \item The court would apply the substantial certainty doctrine if it can be proven that John intentionally removed the safety barrier.
    \item The court would not apply the substantial certainty doctrine because John did not intend for Mike to fall into the pit.
    \item The court would not apply the substantial certainty doctrine because Mike should have been more careful.
\end{enumerate}

The process of expert critiques is iterative if the issue in the original generation is not resolved after one round or a new issue arises. With the renewed choices, the answer generated by LLM becomes:
\begin{quote}
    The right answer is \textbf{(A)}. The substantial certainty doctrine could be applied if it can be proven that John knew with substantial certainty that Mike would fall into the pit. Even though John did not intend for Mike to fall, if he was aware of the high risk of such an outcome when he removed the safety barrier and did not warn Mike, this aligns with the substantial certainty doctrine, which assumes intent even if the actor did not intend the result, but knew with substantial certainty the effect would occur as a result of his action.
\end{quote}

\section{Automatic Complexity Metrics}
\label{appendix:complexity}

Here are the details about the complexity metrics. To tokenize\footnote{We use the ``en\_core\_web\_sm'' spaCy model for tokenization. See \url{https://spacy.io/}.} the stories and definitions, we use:

\paragraph{Legal Vocabulary List (LVL) occurrences} Inspired by \citet{gardner2014new}'s Academic Vocabulary List (AVL), LVL incorporates a list of legalese from the Glossary of Legal Terms from the official US Court website\footnote{\url{https://www.uscourts.gov/glossary}} and the Open Legal Dictionary Project\footnote{\url{https://openlegaldictionary.com/}}. To assess the level of legal rigor in Wikipedia definitions, we quantify the proportion of LVL words employed in each definition.

\paragraph{Top 1000 Most Common Words out-of-vocabulary (Top1K)} The popular book ``Thing Explainer'' utilizes a vocabulary constrained to the 1,000 most frequent English words based on Wiktionary's contemporary fiction frequency list \cite{munroe2015thing}. To assess the simplicity and accessibility of generated definitions, we calculate the proportion of words that are from the top 1,000 words employed in the book.

\paragraph{Function words} In health communication, the use of function words such as prepositions, auxiliary verbs, and question words is positively associated with both perceived and actual readability \cite{leroy2008evaluating,leroy2010influence}. \citet{august-etal-2022-generating} has also applied this to science communication.

\paragraph{Sentence length} Sentence length is a widely used metric for assessing document-level complexity and is incorporated into numerous classic readability measures \cite{pitler-nenkova-2008-revisiting,feng2010comparison}. We only pick concepts whose definitions have 100-200 words and limit the story to 500 words. Therefore, We hypothesize that generated stories will be associated with less complex language due to elaborative simplification, a technique that involves explaining complex terms to facilitate comprehension \cite{srikanth-li-2021-elaborative}.

\paragraph{Language model perplexity} Language model perplexity has demonstrated a positive correlation with perceived and actual reading difficulty \cite{pitler-nenkova-2008-revisiting,collins2014computational}. To assess the complexity of our generated stories, we utilize the GPT model to calculate language model perplexity, considering its training on common English rather than scientific text.

\paragraph{Flesch-Kincaid grade level} This readability score (FK)\footnote{We use the Readability library to compute FK scores. See \url{https://github.com/andreasvc/readability/}.} is derived from straightforward calculations based on sentence length, word length, and syllable counts \cite{kincaid1975derivation}. While studies have shown varying degrees of effectiveness for the FK score in predicting readability in scientific or medical documents \cite{leroy2008evaluating}, it remains a widely used and standardized measure of text complexity \cite{redmiles-etal-2019-comparing}.

\section{Human Evaluation}
\label{appendix:human_prolific}

\subsection{Evaluation Criteria}

\begin{figure}[ht!]
    \centering
    \includegraphics[width=0.9\linewidth]{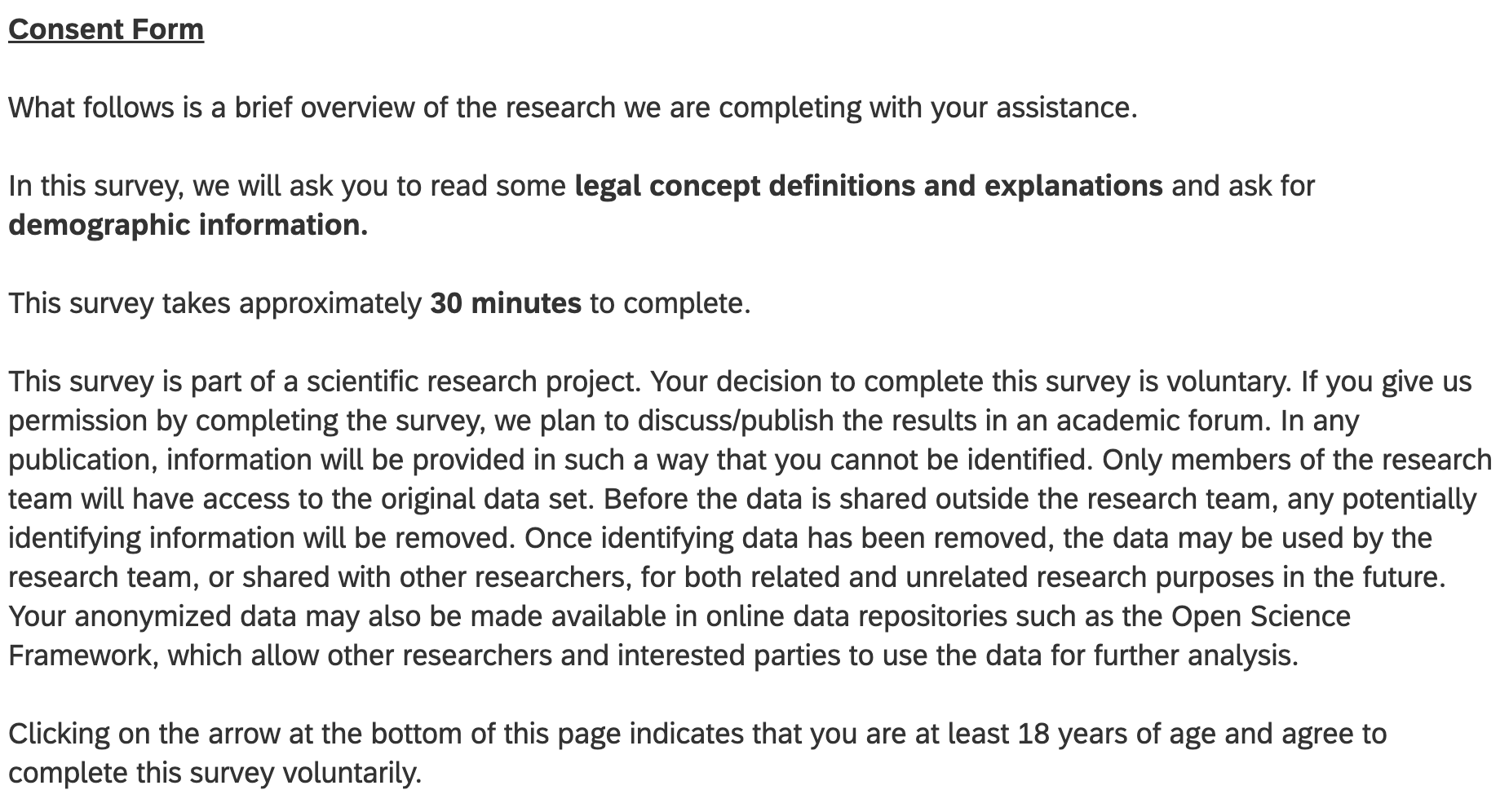}
    \caption{Consent form on Prolific.}
    \label{fig:consent}
\end{figure}

\begin{figure}[ht!]
    \centering
    \includegraphics[width=0.9\linewidth]{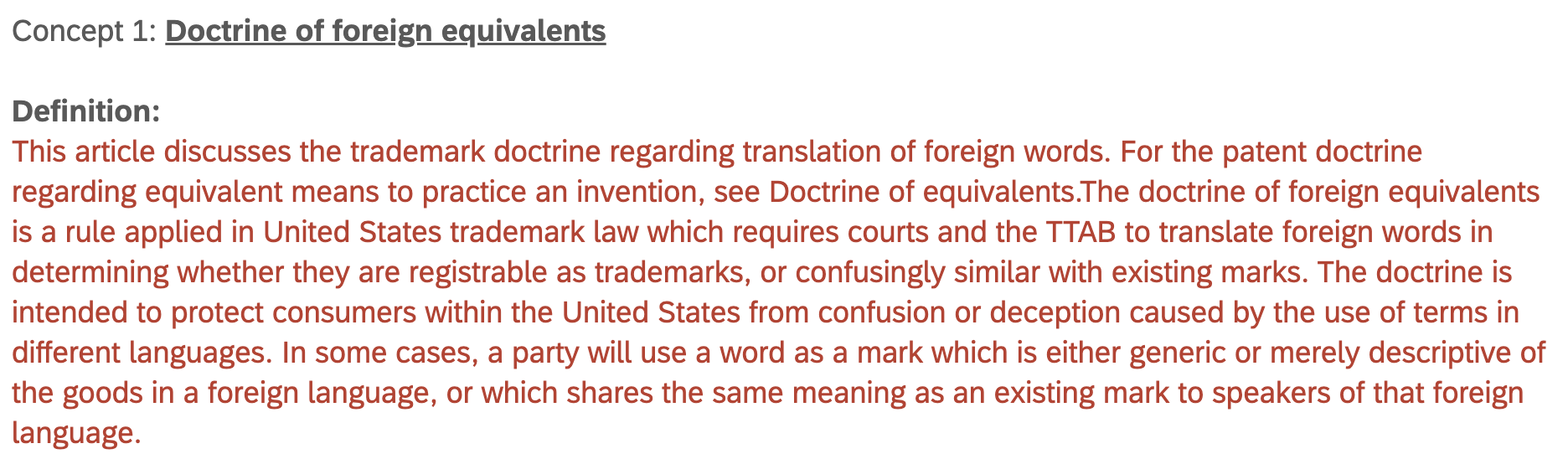}
    \caption{Concept definition example.}
    \label{fig:definition}
\end{figure}

\begin{figure}[ht!]
    \centering
    \includegraphics[width=0.85\linewidth]{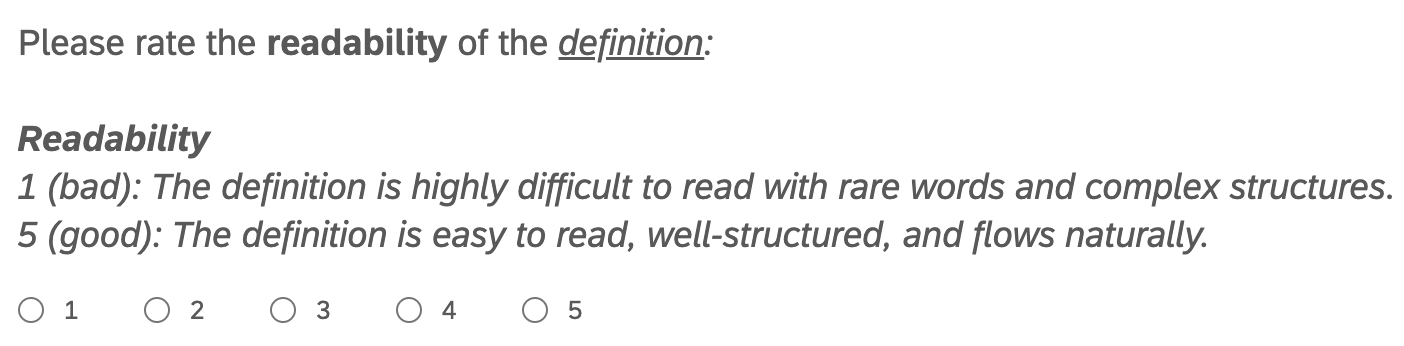}
    \caption{Readability of definition (RoD) question.}
    \label{fig:rod}
\end{figure}

\begin{figure}[ht!]
    \centering
    \includegraphics[width=0.85\linewidth]{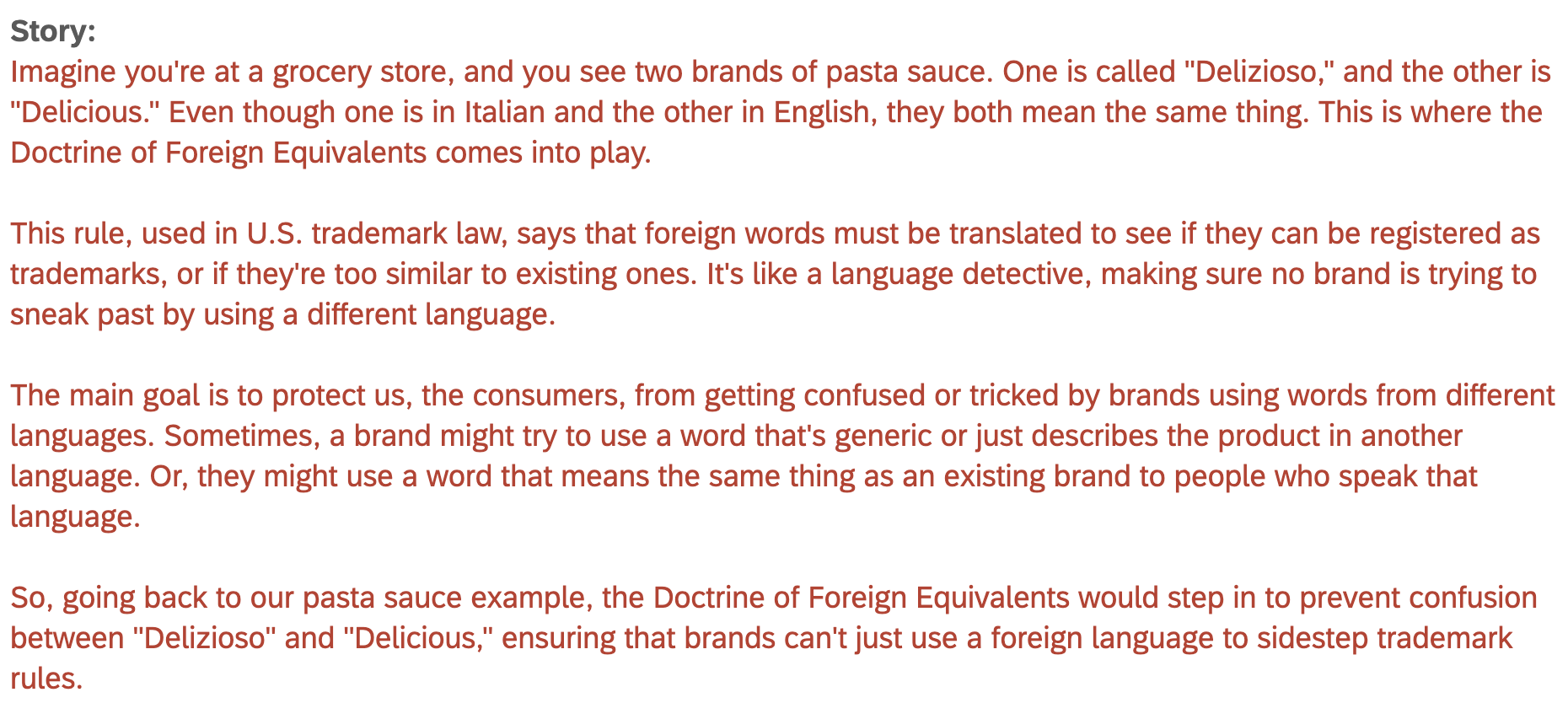}
    \caption{Story example.}
    \label{fig:story}
\end{figure}

\begin{figure}[ht!]
    \centering
    \includegraphics[width=0.85\linewidth]{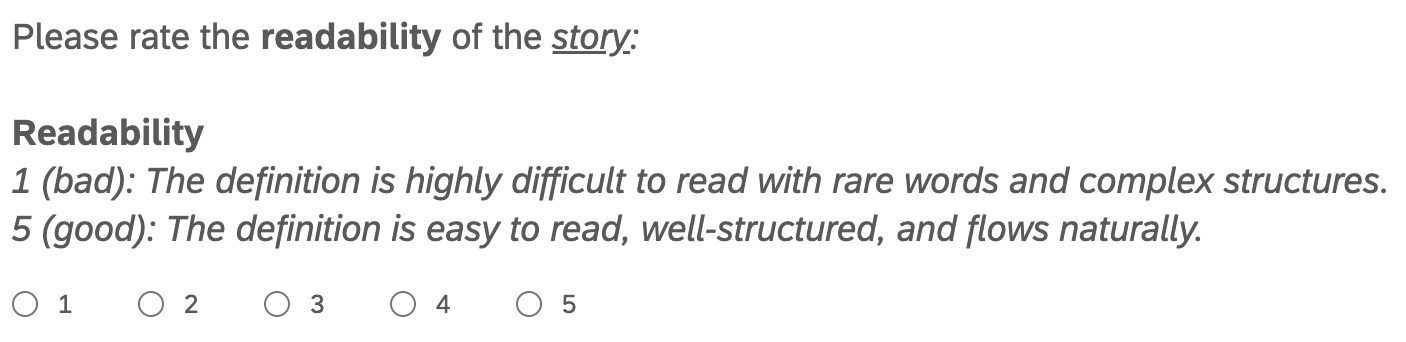}
    \caption{Readability of story (RoS)  question.}
    \label{fig:ros}
\end{figure}

\begin{figure}[ht!]
    \centering
    \includegraphics[width=0.85\linewidth]{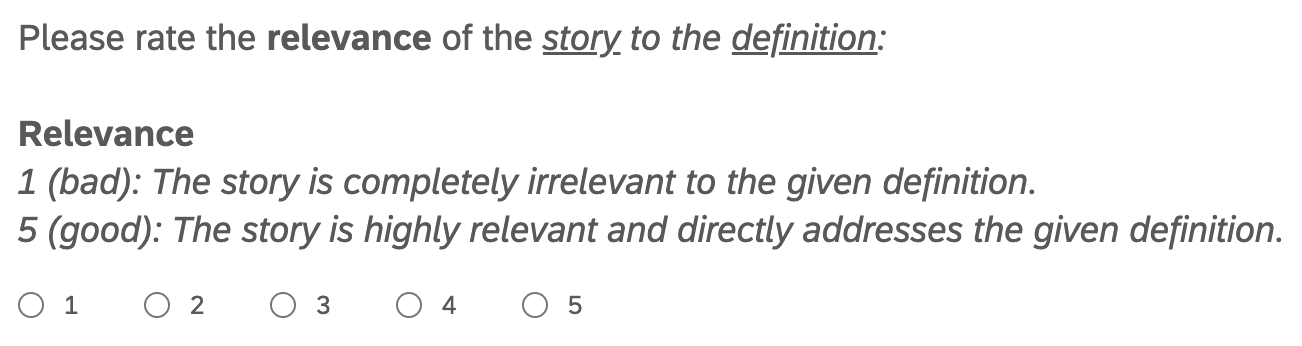}
    \caption{Relevance question.}
    \label{fig:relevance}
\end{figure}

\begin{figure}[ht!]
    \centering
    \includegraphics[width=0.9\linewidth]{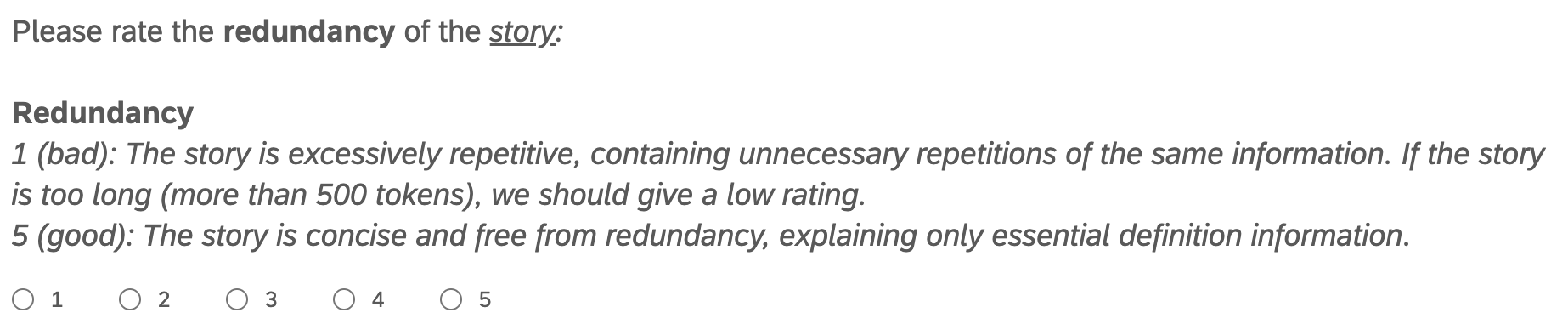}
    \caption{Redundancy question.}
    \label{fig:redundancy}
\end{figure}

\begin{figure}[ht!]
    \centering
    \includegraphics[width=0.9\linewidth]{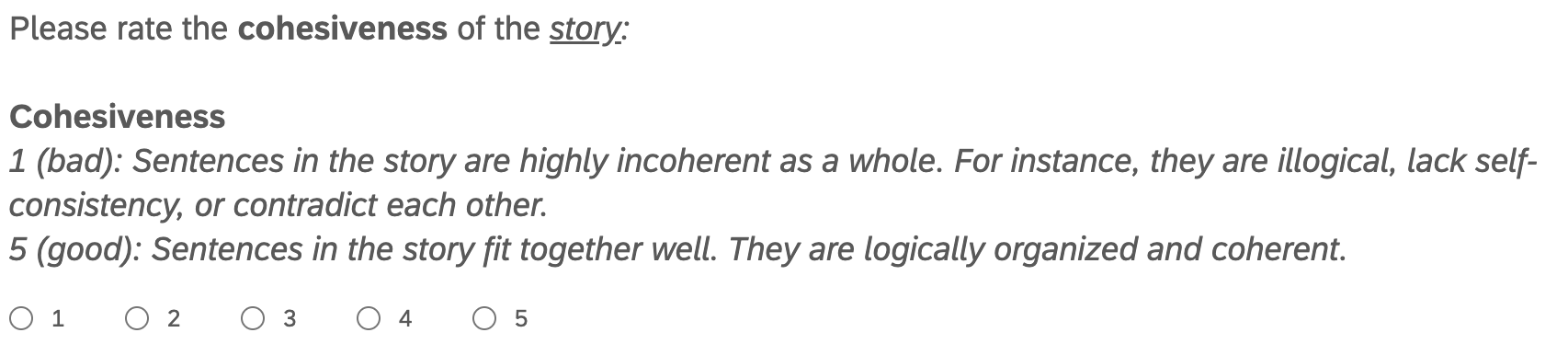}
    \caption{Cohesiveness question.}
    \label{fig:cohesiveness}
\end{figure}

\begin{figure}[ht!]
    \centering
    \includegraphics[width=0.9\linewidth]{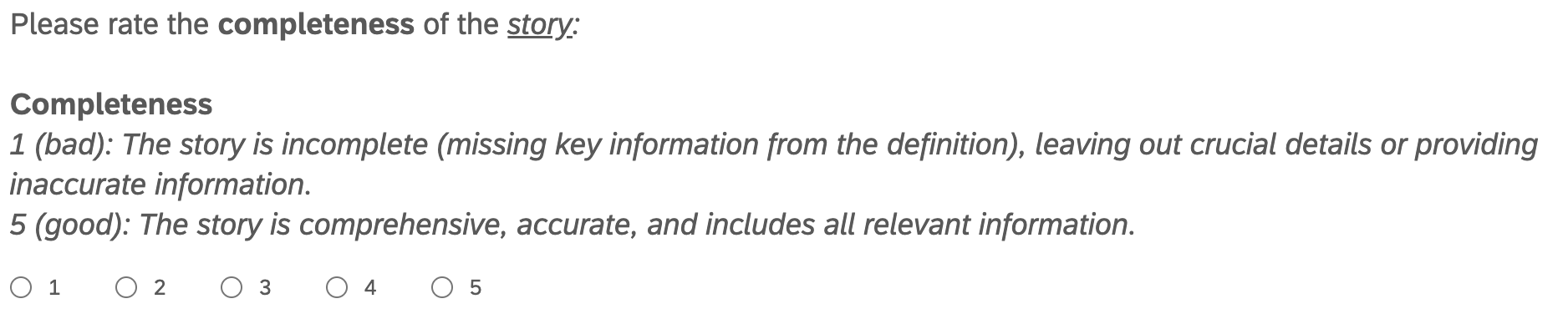}
    \caption{Completeness question.}
    \label{fig:completeness}
\end{figure}

\begin{figure}[ht!]
    \centering
    \includegraphics[width=0.9\linewidth]{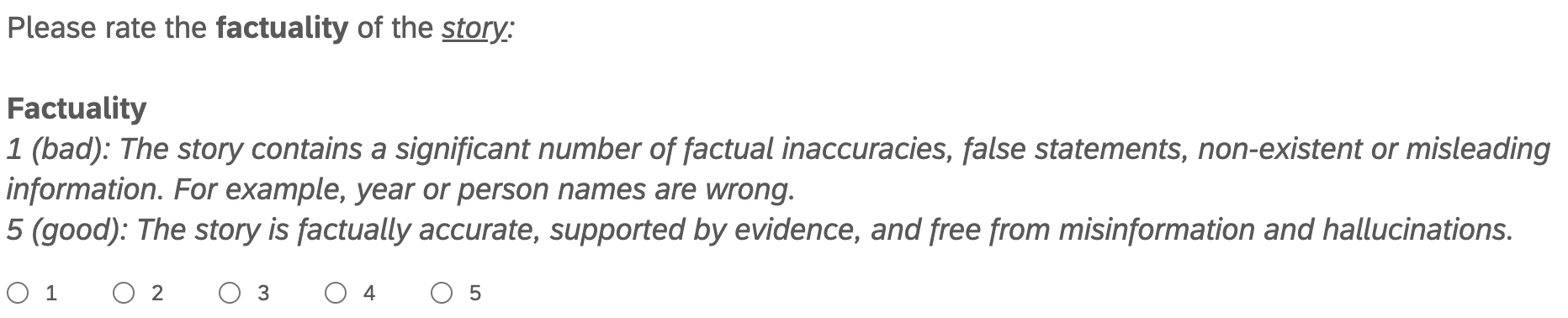}
    \caption{Factuality question.}
    \label{fig:factuality}
\end{figure}

\begin{figure}[ht!]
    \centering
    \includegraphics[width=0.85\linewidth]{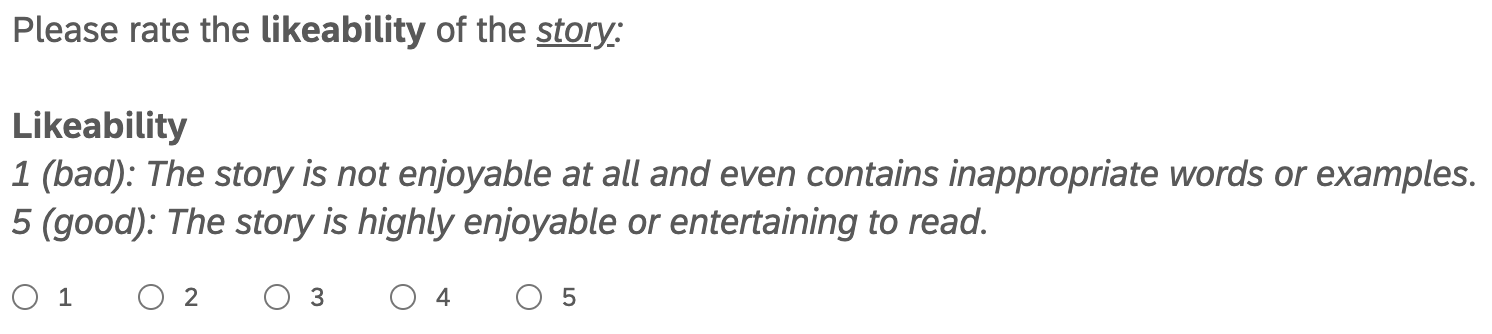}
    \caption{Likeability question.}
    \label{fig:likeability}
\end{figure}

\begin{figure}[ht!]
    \centering
    \includegraphics[width=0.85\linewidth]{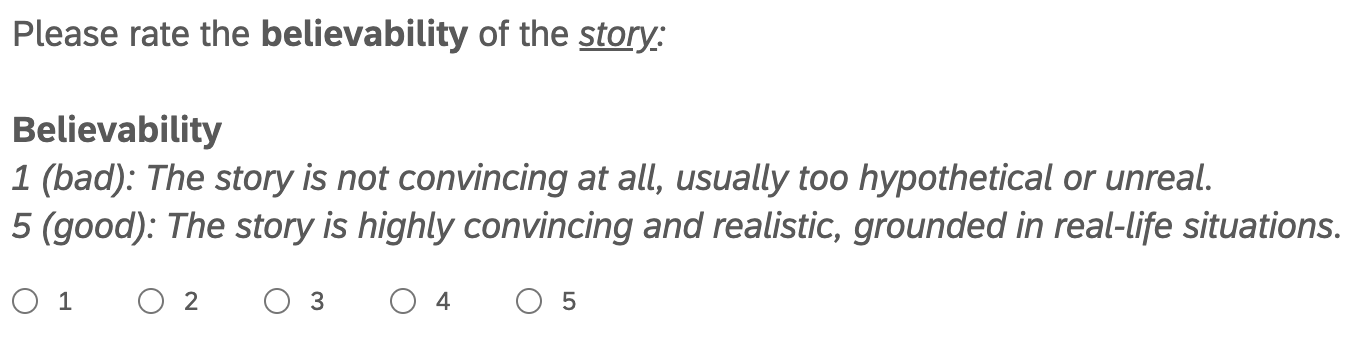}
    \caption{Believability question.}
    \label{fig:believability}
\end{figure}

Details about the Prolific human evaluation, including the survey questions and their demographic details. We recruit Prolific workers who are native English speakers from United States or United Kingdom who have studied law with an approval rate between 99 and 100. Here we include the screenshots for Prolific data evaluation questions. Each batch of annotation is sent to three annotators, which contains 5 legal concepts with their stories and three generated questions. We first present the annotator with a consent form before they proceed (see Figure \ref{fig:consent}). Annotators first look at a concept definition (Figure \ref{fig:definition}) and judge the \textbf{Readability of Definition (RoD)}: whether the concept definition is easy to read, well-structured, and flows naturally (Figure \ref{fig:rod}). Afterward, we ask them see the story (Figure \ref{fig:story}) and judge the \textbf{Readability of Story (RoS)}: whether the story is easy to read, well-structured, and flows naturally (Figure \ref{fig:ros}). Later, they are asked to further evaluate the story along seven dimensions:
\begin{itemize}
    \item \textbf{Relevance}: whether the story is highly relevant and directly addresses the given concept definition. (Figure \ref{fig:relevance})
    \item \textbf{Redundancy}: whether the story is concise and free from unneeded content, explaining only essential definition information. (Figure \ref{fig:redundancy})
    \item \textbf{Cohesiveness}: whether sentences in the story fit together well. (Figure \ref{fig:cohesiveness})
    \item \textbf{Completeness}: whether the story is comprehensive, accurate, and includes all relevant information. (Figure \ref{fig:completeness})
    \item \textbf{Factuality}: whether the story is factually accurate, supported by empirical evidence, and free from misinformation and hallucinations. (Figure \ref{fig:factuality})
    \item \textbf{Likeability}: whether the story is highly enjoyable or entertaining to read. (Figure \ref{fig:likeability})
    \item \textbf{Believability}: whether the story is plausible and internally consistent. (Figure \ref{fig:believability})
\end{itemize}

\begin{figure}[ht!]
    \centering
    \includegraphics[width=0.9\linewidth]{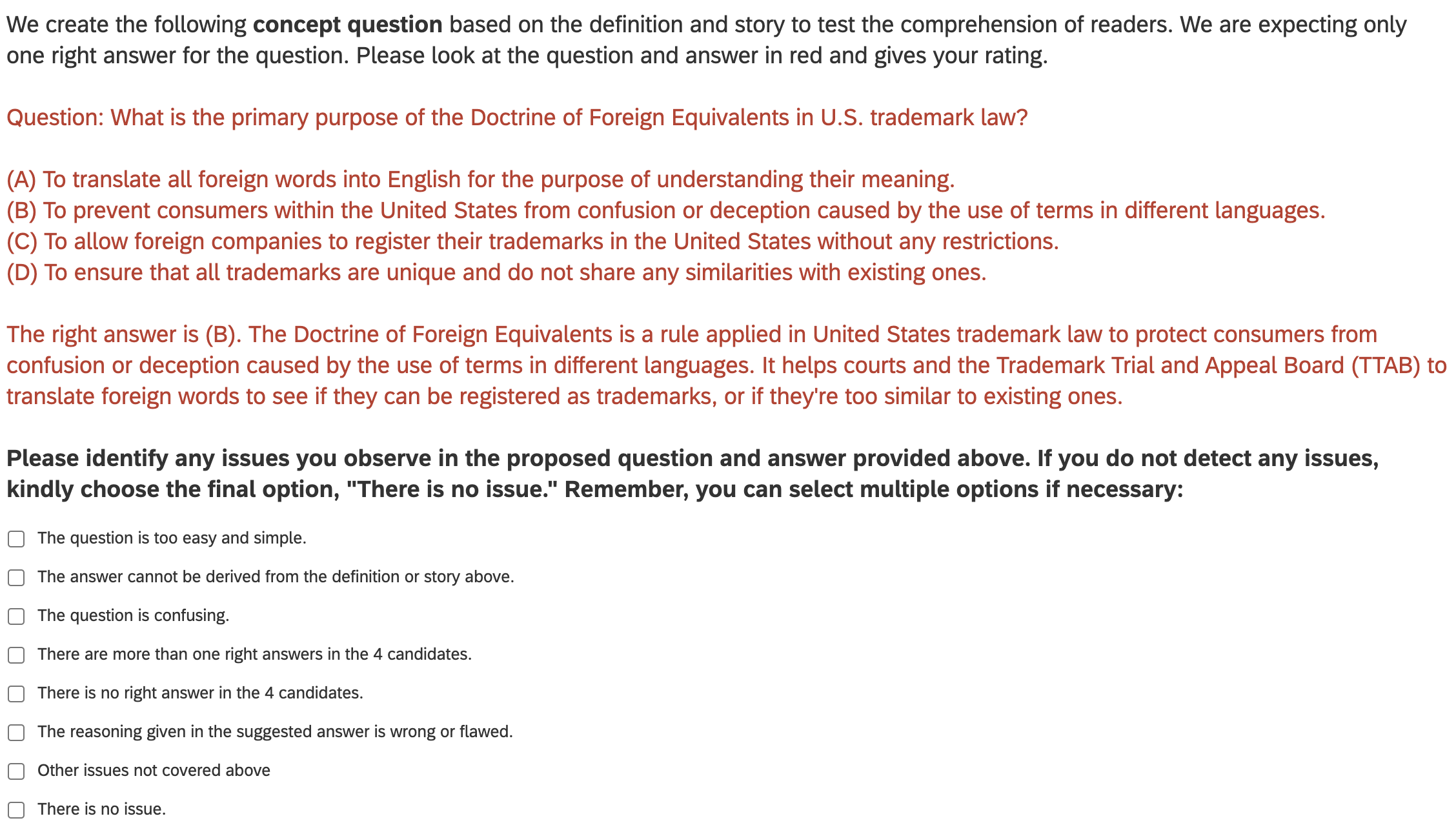}
    \caption{Question example.}
    \label{fig:question}
\end{figure}

\begin{figure}[ht!]
    \centering
    \includegraphics[width=0.9\linewidth]{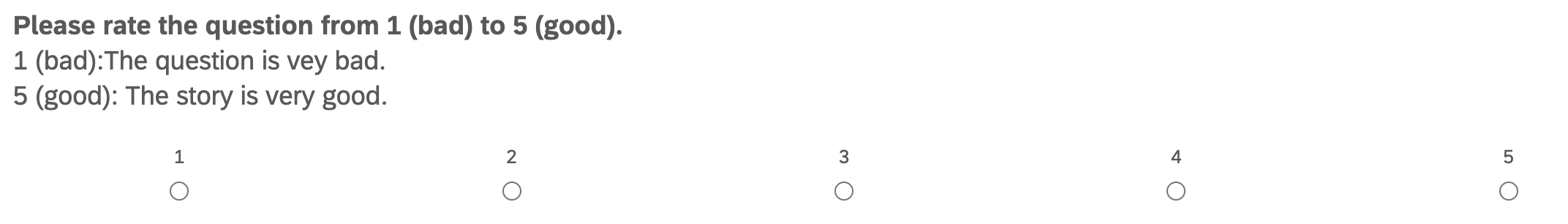}
    \caption{Rating question for the quiz.}
    \label{fig:question_rating}
\end{figure}

After evaluating the story, these annotators are then asked to evaluate the three generated questions along with the suggested answer from LLMs. Here is one example of such question (Figire \ref{fig:question}) and the rating about the question (Figure \ref{fig:question_rating}).

\subsection{Annotator Demographics}
\label{appendix:human_demographics}

\begin{figure}[h]
    \centering
    \begin{subfigure}[b]{0.3\textwidth}
        \includegraphics[width=\textwidth]{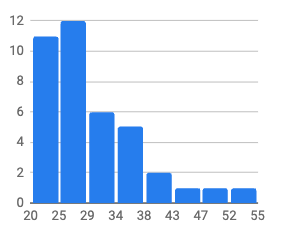}
        \caption{Age}
    \end{subfigure}
    \hfill
    \begin{subfigure}[b]{0.25\textwidth}
        \includegraphics[width=\textwidth]{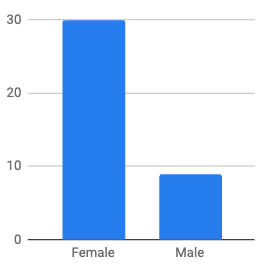}
        \caption{Sex}
    \end{subfigure}
    \hfill
    \begin{subfigure}[b]{0.25\textwidth}
        \includegraphics[width=\textwidth]{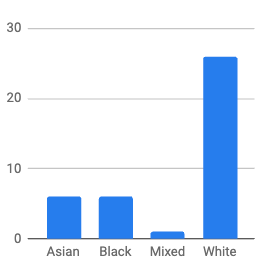}
        \caption{Ethnicity}
    \end{subfigure}
    \caption{Distribution of age, sex, and ethnicity among the 39 Prolific annotators who evaluate the stories.}
    \label{fig:demographics}
\end{figure}

We also include the demographics of 39 unique participants who contribute to evaluate the stories and questions. 26 participants are from UK and 13 from US. We show the distribution of age, sex, and ethnicity in the Figure \ref{fig:demographics}.

\section{RCT Experiment Details}
\label{appendix:rct_details}

\subsection{RCT Procedure}
We recruit participants on Prolific with the following criteria: (a) have little to no law backgrounds, (b) have a bachelor's degrees as the person's highest degree, (c) lives or work in the North American region, (d) have an approval rate between 98 and 100. The criteria are set to have lower and upper bounds for language proficiency and background to limit the variance that can be sufficiently analyzed with our small sample size. After the pre-survey, the participants were randomized into two conditions based on age, gender, and educational background. We then deploy the tasks as batches of 5 concepts and recruit 15 to 20 people to complete the tasks; each person can only complete the same concept once. Due to the difference in completion levels, we have varying responses for each concept (ranging from 16 to 20). 

\begin{figure}[ht!]
    \centering
    \includegraphics[width=0.9\linewidth]{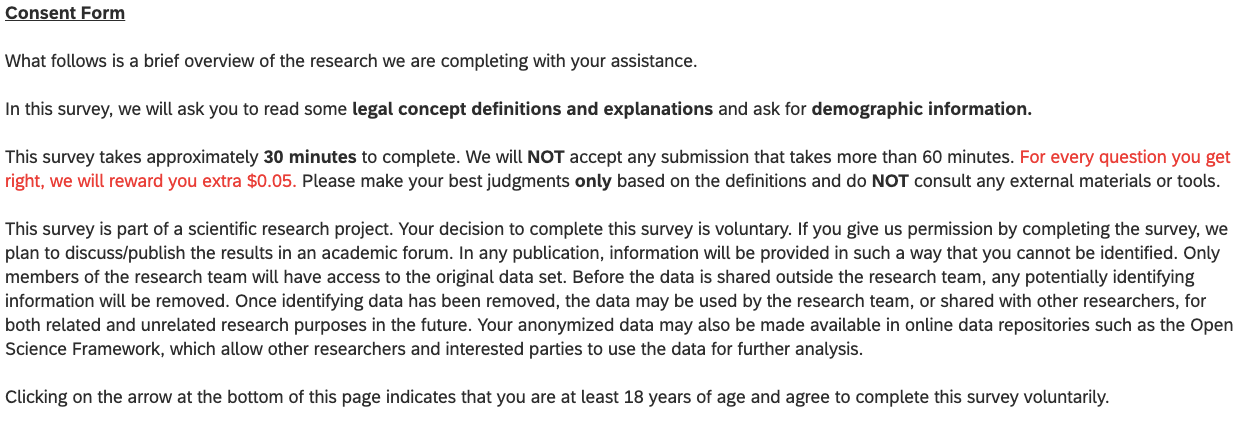}
    \caption{Consent form in the RCT.}
    \label{fig:rct_consent}
\end{figure}

\begin{figure}[ht!]
    \centering
    \includegraphics[width=0.85\linewidth]{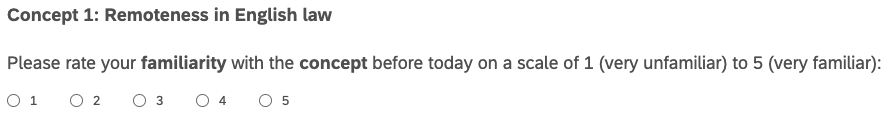}
    \caption{Concept familiarity in the RCT.}
    \label{fig:rct_fam}
\end{figure}

\begin{figure}[ht!]
    \centering
    \includegraphics[width=0.9\linewidth]{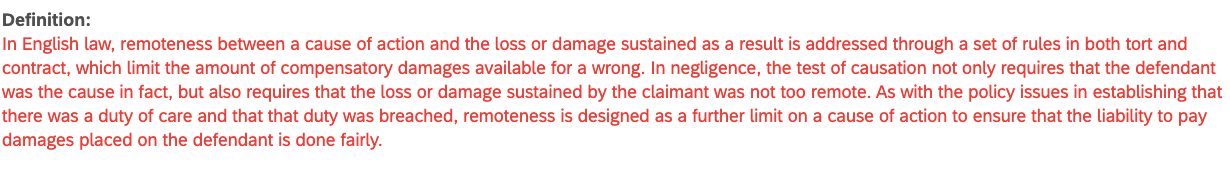}
    \caption{Concept Definition example in the RCT.}
    \label{fig:rct_definition}
\end{figure}

\begin{figure}[ht!]
    \centering
    \includegraphics[width=0.9\linewidth]{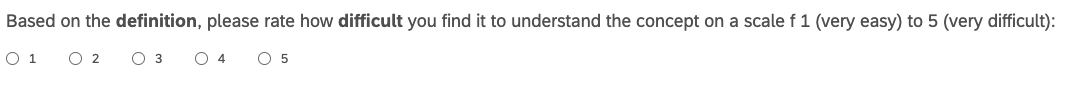}
    \caption{Perceived difficulty in the RCT.}
    \label{fig:rct_diff}
\end{figure}

\begin{figure}[ht!]
    \centering
    \includegraphics[width=0.9\linewidth]{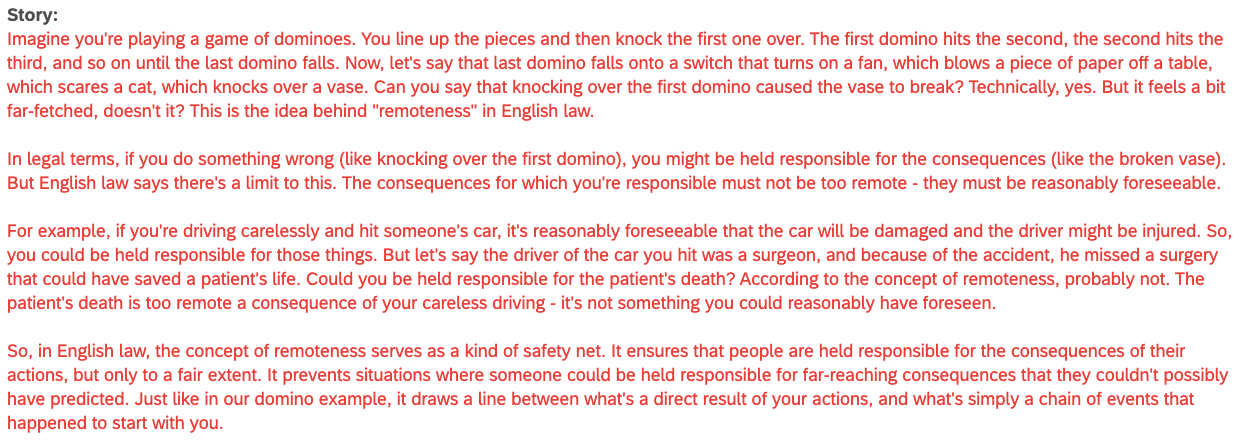}
    \caption{Story example in the RCT.}
    \label{fig:rct_story}
\end{figure}

\begin{figure}[ht!]
    \centering
    \includegraphics[width=0.9\linewidth]{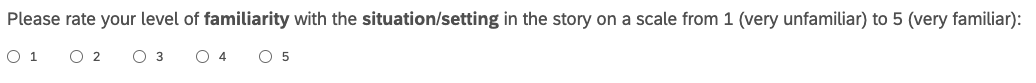}
    \caption{Familiarity with the story setting in the RCT.}
    \label{fig:rct_fam_story}
\end{figure}

\begin{figure}[ht!]
    \centering
    \includegraphics[width=0.9\linewidth]{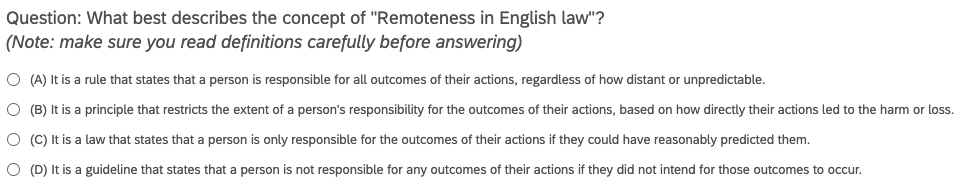}
    \caption{Concept question example in the RCT.}
    \label{fig:rct_q1}
\end{figure}

\begin{figure}[ht!]
    \centering
    \includegraphics[width=0.9\linewidth]{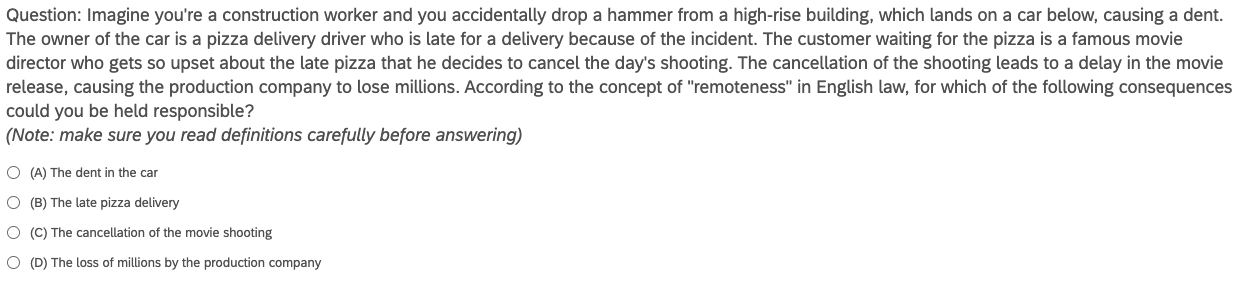}
    \caption{Prediction question example in the RCT.}
    \label{fig:rct_q2}
\end{figure}

\begin{figure}[ht!]
    \centering
    \includegraphics[width=0.9\linewidth]{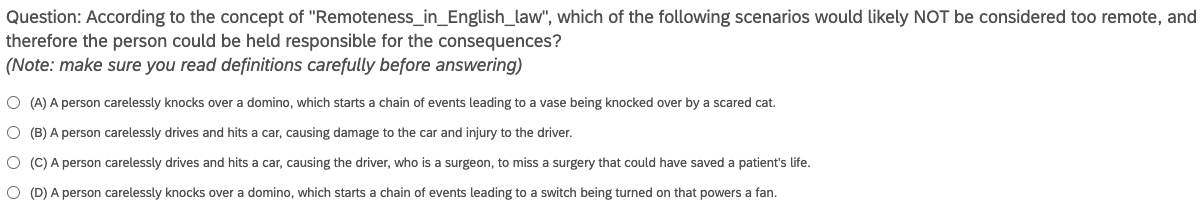}
    \caption{Limitation question example in the RCT.}
    \label{fig:rct_q3}
\end{figure}

In the procedure of the RCT experiment, we begin with a consent form to ensure participants understand the task and how we want to use the data shown in Figure~\ref{fig:rct_consent}. In order to encourage participants to complete the questions accurately, we offer a reward of $\$0.05$ for each question that is answered correctly within 60 minutes. Afterwards, we present the participants with the concept and ask them to evaluate their familiarity of the concept (Figure \ref{fig:rct_fam}). After reading the concept definition (Figure \ref{fig:rct_definition}, the participant need to evaluate the perceived difficulty of the concept (Figure \ref{fig:rct_diff}). If a participant is assigned to the treatment group, the participant will read a story (Figure \ref{fig:rct_story}) and evaluate the familiarity of the story setting to the person (Figure \ref{fig:rct_fam_story}). Otherwise, the participant will skip this story reading part. Afterwards, the participant will answer three questions including the concept qustion (Figure \ref{fig:rct_q1}), the prediction question (Figure \ref{fig:rct_q2}), and the limitation question (Figure \ref{fig:rct_q3}). At the end of the study, all participants will be asked if they are interested in learning more about law and legal knowledge.

\begin{figure}[ht!]
    \centering
    \includegraphics[width=0.9\linewidth]{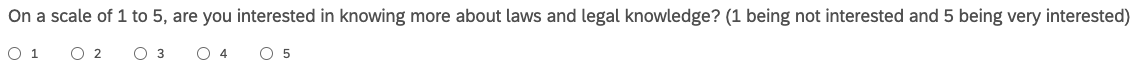}
    \caption{Law interst near the end of the RCT.}
    \label{fig:rct_interest}
\end{figure}

\subsection{Participant Demographics}
\begin{figure}[h]
    \centering
    \begin{subfigure}[b]{0.3\textwidth}
        \includegraphics[width=\textwidth]{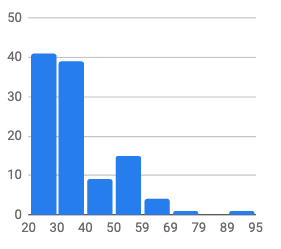}
        \caption{Age}
    \end{subfigure}
    \hfill
    \begin{subfigure}[b]{0.27\textwidth}
        \includegraphics[width=\textwidth]{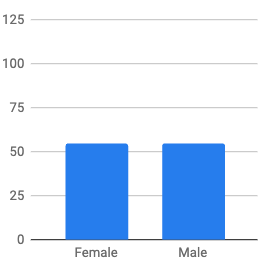}
        \caption{Sex}
    \end{subfigure}
    \hfill
    \begin{subfigure}[b]{0.27\textwidth}
        \includegraphics[width=\textwidth]{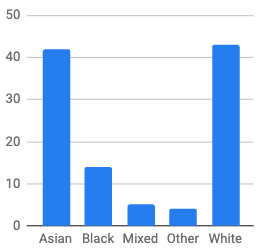}
        \caption{Ethnicity}
    \end{subfigure}
    \caption{Distribution of age, sex, and ethnicity among the 110 RCT participants from Prolific.}
    \label{fig:rct_demographics}
\end{figure}

There are 65 native respondents and 71 non-native respondents in the study. Some respondents participated in two batches and some participated in just one batch. Out of 136 respondents, there are 117 unique participants in our RCT experiments and we are able to collect the demographics of 110 participants. 35 are from Canada and 75 are from the United States. 70 report their first language as English and 40 report their first language as other languages. We include the distribution of age, sex, and ethnicity in the Figure \ref{fig:rct_demographics}.

\subsection{Individual Accuracy}
\label{appendix:individual_accuracy}
We provide the individual level accuracy in addition to the result presented in the main paper as a reference. In the human-subject study, we distribute the 10 concepts in 2 batches (5 concepts in each batch) on Prolific to prevent overloading the participants and losing their engagement. It is worth noting that it takes 20 to 25 minutes to complete the study with 5 concepts. In the following Table \ref{tab:rct_individual_accuracy_1} and \ref{tab:rct_individual_accuracy_2}, we report the individual accuracy by batches.

\setlength{\extrarowheight}{2pt}
\begin{table}[th!]
    \centering
    \footnotesize
    \begin{tabular}{cccc}
        {\textbf{Condition}}  & 
        {\textbf{ConceptQ}}  &
        {\textbf{PredictionQ}} & 
        {\textbf{LimitationQ}} \\
         \Xhline{3\arrayrulewidth}
        \multicolumn{4}{c}{\textit{Native Speakers (Individual Accuracy - Batch 1)}} \\ 
        Definition & $97.65 \pm 9.41$ & $78.82 \pm 24.22$ & $80.00 \pm 24.73$ \\
        Def. + Story & $91.25 \pm 19.96$ & $77.50 \pm 29.05$ & $83.75 \pm 24.71$ \\
        \Xhline{3\arrayrulewidth}
        \multicolumn{4}{c}{\textit{Non-native Speakers (Individual Accuracy - Batch 1)}} \\ 
        Definition & $89.52 \pm 15.88$ & $73.33 \pm 24.94$ & $73.33 \pm 24.16$ \\
        Def. + Story & $92.22 \pm 13.56$ & $84.44 \pm 17.07$ & $92.22 \pm 15.11$ \\
    \end{tabular}
    \caption{Average and Standard Deviation of Individual Level Accuracy for Batch 1}
    \label{tab:rct_individual_accuracy_1}
\end{table}

\setlength{\extrarowheight}{2pt}
\begin{table}[th!]
    \centering
    \footnotesize
    \begin{tabular}{cccc}
        {\textbf{Condition}}  & 
        {\textbf{ConceptQ}}  &
        {\textbf{PredictionQ}} & 
        {\textbf{LimitationQ}} \\
         \Xhline{3\arrayrulewidth}
        \multicolumn{4}{c}{\textit{Native Speakers (Individual Accuracy - Batch 2)}} \\ 
        Definition & $88.75 \pm 12.18$ & $78.75 \pm 16.54$ & $75.00 \pm 18.03$ \\
        Def. + Story & $90.00 \pm 17.32$ & $71.25 \pm 22.33$ & $85.00 \pm 22.91$ \\
        \Xhline{3\arrayrulewidth}
        \multicolumn{4}{c}{\textit{Non-native Speakers (Individual Accuracy - Batch 2)}} \\ 
        Definition & $88.75 \pm 17.28$ & $70.00 \pm 25.50$ & $62.50 \pm 27.27$ \\
        Def. + Story & $90.00 \pm 12.25$ & $78.75 \pm 17.98$ & $76.25 \pm 17.63$ \\
    \end{tabular}
    \caption{Average and Standard Deviation of Individual Level Accuracy for Batch 2}
    \label{tab:rct_individual_accuracy_2}
\end{table}

\end{document}